\documentclass[11pt]{article}
\pdfoutput=1

\usepackage[utf8]{inputenc}
\usepackage{amsmath,amssymb,amsthm}
\usepackage{array}
\usepackage{authblk}
\usepackage[english]{babel}
\usepackage{bm}
\usepackage{booktabs}
\usepackage{caption}
\usepackage{courier}
\usepackage{csquotes}
\usepackage{float}
\usepackage[top=1.0in,bottom=1.0in,left=1.1in,right=1.1in]{geometry}
\usepackage{graphicx}
\usepackage{mathtools}
\usepackage{mdframed}
\usepackage[square,numbers]{natbib}
\usepackage{subcaption}
\usepackage{subfiles}
\usepackage{xcolor}

\graphicspath{ {images/} }

\title{
    \vspace{-5em}
    \begin{mdframed}[nobreak=true]
        \normalsize{
            \noindent
            This manuscript is published in Neural Networks. Please cite it as:\\
            Mateusz Buda, Atsuto Maki, and Maciej A Mazurowski. A systematic study of the class imbalance problem in convolutional neural networks. Neural Networks, 106:249–259, 2018.
        }
    \end{mdframed}
    \vspace{2.5em}
    A systematic study of the class imbalance problem \\in convolutional neural networks\footnote{\textregistered~2018. This manuscript version is made available under the CC-BY-NC-ND 4.0 license.}
}
\date{}

\author[1, 2]{Mateusz Buda}
\author[2]{Atsuto Maki}
\author[1, 3]{Maciej A. Mazurowski}

\affil[ ]{\footnotesize }
\affil[1]{\footnotesize Department of Radiology, Duke University School of Medicine, Durham, NC, USA}
\affil[2]{\footnotesize Royal Institute of Technology (KTH), Stockholm, Sweden}
\affil[3]{\footnotesize Department of Electrical and Computer Engineering, Duke University, Durham, NC, USA}
\affil[ ]{\footnotesize }
\affil[ ]{\texttt{\{buda, atsuto\}@kth.se} \hspace{1.0em} \texttt{maciej.mazurowski@duke.edu}}

\begin{document}

{\let\newpage\relax\maketitle}
\vspace{-2em}

\begin{abstract}

In this study, we systematically investigate the impact of class imbalance on classification performance of convolutional neural networks (CNNs) and compare frequently used methods to address the issue.
Class imbalance is a common problem that has been comprehensively studied in classical machine learning, yet very limited systematic research is available in the context of deep learning.
In our study, we use three benchmark datasets of increasing complexity, MNIST, CIFAR\nobreakdash-10 and ImageNet, to investigate the effects of imbalance on classification and perform an extensive comparison of several methods to address the issue: oversampling, undersampling, two\nobreakdash-phase training, and thresholding that compensates for prior class probabilities.
Our main evaluation metric is area under the receiver operating characteristic curve (ROC AUC) adjusted to multi-class tasks since overall accuracy metric is associated with notable difficulties in the context of imbalanced data.
Based on results from our experiments we conclude that
(i)~the effect of class imbalance on classification performance is detrimental;
(ii)~the method of addressing class imbalance that emerged as dominant in almost all analyzed scenarios was oversampling;
(iii)~oversampling should be applied to the level that completely eliminates the imbalance, whereas the optimal undersampling ratio depends on the extent of imbalance;
(iv)~as opposed to some classical machine learning models, oversampling does not cause overfitting of CNNs;
(v)~thresholding should be applied to compensate for prior class probabilities when overall number of properly classified cases is of interest.

\vspace{0.4em}

\end{abstract}

\section{Introduction}
\label{sec:introduction}

Convolutional neural networks (CNNs) are gaining significance in a number of machine learning application domains and are currently contributing to the state~of~the~art in the field of computer vision, which includes tasks such as object detection, image classification, and segmentation.
They are also widely used in natural language processing or speech recognition where they are replacing or improving classical machine learning models~\cite{gu2015recent}.
CNNs integrate automatic feature extraction and discriminative classifier in one model, which is the main difference between them and traditional machine learning techniques.
This property allows CNNs to learn hierarchical representations~\cite{zeiler2014visualizing}.
The standard CNN is built with fully connected layers and a number of blocks consisting of convolutions, activation function layer and max pooling~\cite{lecun1989backpropagation, krizhevsky2012imagenet, simonyan2014very}.
The complex nature of CNNs requires a significant computational power for training and evaluation of the networks, which is addressed with the help of modern graphical processing units (GPUs).

A common problem in real life applications of deep learning based classifiers is that some classes have a significantly higher number of examples in the training set than other classes.
This difference is referred to as class imbalance.
There are plenty of examples in domains like computer vision~\cite{van2017inaturalist, xiao2010sun, johnson2013hybrid, kubat1998machine, beijbom2012automated}, medical diagnosis~\cite{grzymala2004approach, mac2002problem}, fraud detection~\cite{chan1998toward} and others~\cite{radivojac2004classification, cardie1997improving, haixiang2016learning} where this issue is highly significant and the frequency of one class (e.g., cancer) can be 1000 times less than another class (e.g., healthy patient).
It has been established that class imbalance can have significant detrimental effect on training traditional classifiers~\cite{japkowicz2002class} including multi-layer perceptrons~\cite{mazurowski2008training}.
It affects both convergence during the training phase and generalization of a model on the test set.
While the issue very likely also affects deep learning, no systematic study on the topic is available.

Methods of dealing with imbalance are well studied for classical machine learning models~\cite{chawla2005data, japkowicz2002class, maloof2003learning, mazurowski2008training}.
The most straightforward and common approach is the use of sampling methods.
Those methods operate on the data itself (rather than the model) to increase its balance.
Widely used and proven to be robust is oversampling~\cite{ling1998data}.
Another option is undersampling.
Na\"ive version, called random majority undersampling, simply removes a random portion of examples from majority classes~\cite{japkowicz2002class}.
The issue of class imbalance can be also tackled on the level of the classifier.
In such case, the learning algorithms are modified, e.g. by introducing different weights to misclassification of examples from different classes~\cite{zhou2006training} or explicitly adjusting prior class probabilities~\cite{lawrence1998neural}.

Some previous studies showed results on cost sensitive learning of deep neural networks~\cite{khan2015cost, raj2016towards, chung2015cost}.
New kinds of loss function for neural networks training were also developed~\cite{wang2016training}.
Recently, a new method for CNNs was introduced that trains the network in two-phases in which the network is trained on the balanced data first and then the output layers are fine-tuned~\cite{havaei2017brain}.
While little systematic analysis of imbalance and methods to deal with it is available for deep learning, researchers employ some methods that might be addressing the problem likely based on intuition, some internal tests, and systematic results available for traditional machine learning.
Based on our review of the literature, the method most commonly applied in deep learning is oversampling.

The reminder of this paper is organized as follows.
Section~\ref{sec:related} gives an overview of methods to address the problem of imbalance.
In Section~\ref{sec:experiments} we describe the experimental setup.
It provides details about compared methods, datasets and models used for evaluation.
Then, in Section~\ref{sec:results} we present the results from our experiments and compare methods.
Finally, Section~\ref{sec:conclusions} concludes the paper.

\section{Methods for addressing imbalance}
\label{sec:related}

Methods for addressing class imbalance can be divided into two main categories~\cite{he2009learning}.
The first category is data level methods that operate on training set and change its class distribution.
They aim to alter dataset in order to make standard training algorithms work.
The other category covers classifier (algorithmic) level methods.
These methods keep the training dataset unchanged and adjust training or inference algorithms.
Moreover, methods that combine the two categories are available.
In this section we give an overview of commonly used approaches in both classical machine learning models and deep neural networks.

\subsection{Data level methods}
\paragraph{Oversampling.}
One of the most commonly used method in deep learning~\cite{haixiang2016learning, levi2015age, janowczyk2016deep, jaccard2016detection}.
The basic version of it is called random minority oversampling, which simply replicates randomly selected samples from minority classes.
It has been shown that oversampling is effective, yet it can lead to overfitting~\cite{chawla2002smote, wang2014hybrid}.
A more advanced sampling method that aims to overcome this issue is SMOTE~\cite{chawla2002smote}.
It augments artificial examples created by interpolating neighboring data points.
Some extensions of this technique were proposed, for example focusing only on examples near the boundary between classes~\cite{han2005borderline}.
Another type of oversampling approach uses data preprocessing to perform more informed oversampling.
Cluster-based oversampling first clusters the dataset and then oversamples each cluster separately~\cite{jo2004class}.
This way it reduces both between-class and within-class imbalance.
DataBoost\nobreakdash-IM, on the other hand, identifies difficult examples with boosting preprocessing and uses them to generate synthetic data~\cite{guo2004learning}.
An oversampling approach specific to neural networks optimized with stochastic gradient descent is class-aware sampling~\cite{shen2016relay}.
The main idea is to ensure uniform class distribution of each mini-batch and control the selection of examples from each class.

\paragraph{Undersampling.}
Another popular method~\cite{haixiang2016learning} that results in having the same number of examples in each class.
However, as opposed to oversampling, examples are removed randomly from majority classes until all classes have the same number of examples.
While it might not appear intuitive, there is some evidence that in some situations undersampling can be preferable to oversampling~\cite{drummond2003c4}.
A significant disadvantage of this method is that it discards a portion of available data.
To overcome this shortcoming, some modifications were introduced that more carefully select examples to be removed.
E.g. one-sided selection identifies redundant examples close to the boundary between classes~\cite{kubat1997addressing}.
A more general approach than undersampling is data decontamination that can involve relabeling of some examples~\cite{koplowitz1981relation, barandela2003restricted}.

\subsection{Classifier level methods}

\paragraph{Thresholding.}
Also known as threshold moving or post scaling, adjusts the decision threshold of a classifier.
It is applied in the test phase and involves changing the output class probabilities.
There are many ways in which the network outputs can be adjusted.
In general, the threshold can be set to minimize arbitrary criterion using an optimization algorithm~\cite{lawrence1998neural}.
However, the most basic version simply compensates for prior class probabilities~\cite{richard1991neural}.
These are estimated for each class by its frequency in the imbalanced dataset before sampling is applied.
It was shown that neural networks estimate Bayesian a posteriori probabilities~\cite{richard1991neural}.
That is, for a given datapoint $x$, their output for class $i$ implicitly corresponds to
\begin{align*}
    y_i(x) = p(i|x) = \frac{p(i) \cdot p(x|i)}{p(x)}.
\end{align*}
Therefore, correct class probabilities can be obtained by dividing the network output for each class by its estimated prior probability ${p(i) = \frac{|i|}{\sum_{k}{|k|}}}$, where $|i|$ denotes the number of unique examples in class $i$.

\paragraph{Cost sensitive learning.}
This method assigns different cost to misclassification of examples from different classes~\cite{elkan2001foundations}.
With respect to neural networks it can be implemented in various ways.
One approach is threshold moving~\cite{zhou2006training} or post scaling~\cite{lawrence1998neural} that is applied in the inference phase after the classifier is already trained.
Similar strategy is to adapt the output of the network and also use it in the backward pass of backpropagation algorithm~\cite{kukar1998cost}.
Another adaptation of neural network to be cost sensitive is to modify the learning rate such that higher cost examples contribute more to the update of weights.
And finally we can train the network by minimizing the misclassification cost instead of standard loss function~\cite{kukar1998cost}.
The results of this approach are equivalent to oversampling~\cite{zhou2006training, chung2015cost} described above and therefore this method will not be implemented in our study.

\paragraph{One-class classification.}
In the context of neural networks it is usually called novelty detection.
This is a concept learning technique that recognizes positive instances rather than discriminating between two classes.
Autoencoders used for this purpose are trained to perform autoassociative mapping, i.e. identity function.
Then, the classification of a new example is made based on a reconstruction error between the input and output patterns, e.g. absolute error, squared sum of errors, Euclidean or Mahalanobis distance~\cite{japkowicz1995novelty, japkowicz2000nonlinear, sohn2001novelty}.
This method has proved to work well for extremely high imbalance when classification problem turns into anomaly detection~\cite{lee2006novelty}.

\paragraph{Hybrid of methods.}
This is an approach that combines multiple techniques from one or both abovementioned categories.
Widely used example is ensembling.
It can be viewed as a wrapper to other methods.
\textit{EasyEnsemble} and \textit{BalanceCascade} are methods that train a committee of classifiers on undersampled subsets~\cite{liu2009exploratory}.
SMOTEBoost, on the other hand, is a combination of boosting and SMOTE oversampling~\cite{chawla2003smoteboost}.
Recently introduced and successfully applied to CNN training for brain tumor segmentation, is two-phase training~\cite{havaei2017brain}.
Even though the task was image segmentation, it was approached as a pixel level classification.
The method involves network pre-training on balanced dataset and then fine-tuning the last output layer before softmax on the original, imbalanced data.

\section{Experiments}
\label{sec:experiments}

\subsection{Forms of imbalance}
\label{sec:imbalance}
Class imbalance can take many forms particularly in the context of multiclass classification, which is typical in CNNs.
In some problems only one class might be underrepresented or overrepresented and in other every class will have a different number of examples.
In this study we define and investigate two types of imbalance that we believe are representative of most of the real-world cases.

\begin{figure}[!ht]
    \centering
    \begin{subfigure}[b]{0.31\textwidth}
            \includegraphics[width=\linewidth]{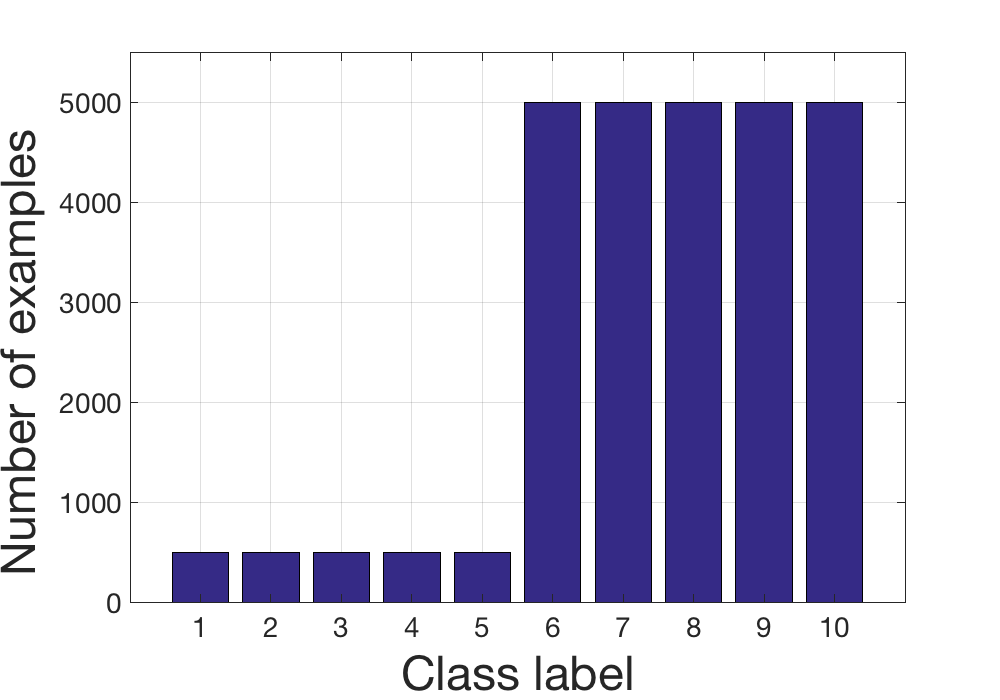}
            \caption{${\rho = 10, \mu = 0.5}$}
            \label{fig:imbalance-step10}
    \end{subfigure}
    \begin{subfigure}[b]{0.31\textwidth}
            \includegraphics[width=\linewidth]{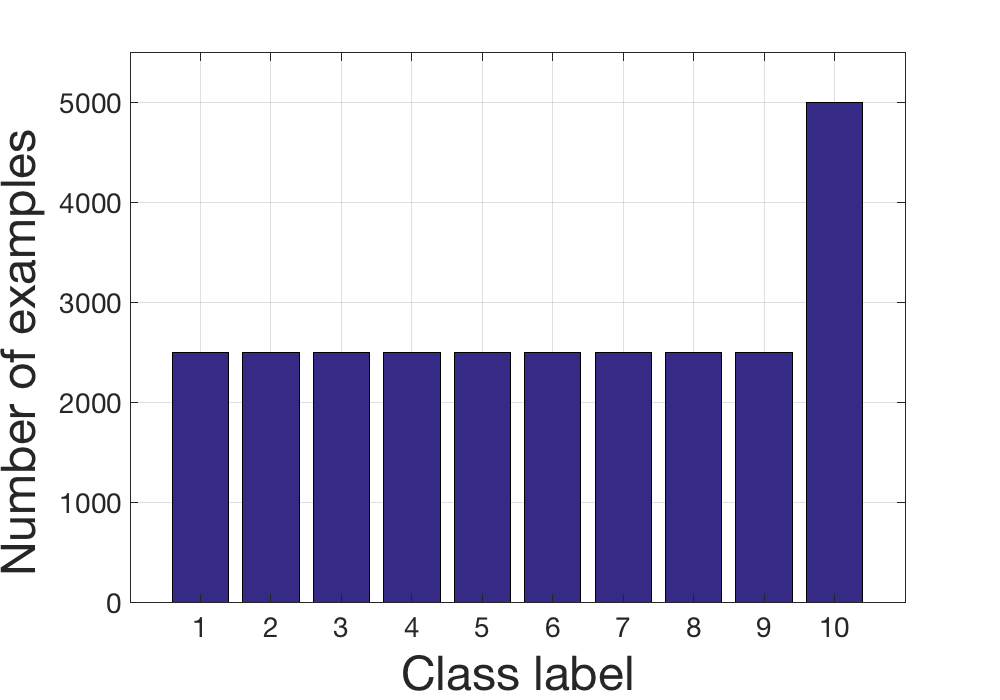}
            \caption{${\rho = 2, \mu = 0.9}$}
            \label{fig:imbalance-step2}
    \end{subfigure}
    \begin{subfigure}[b]{0.31\textwidth}
            \includegraphics[width=\linewidth]{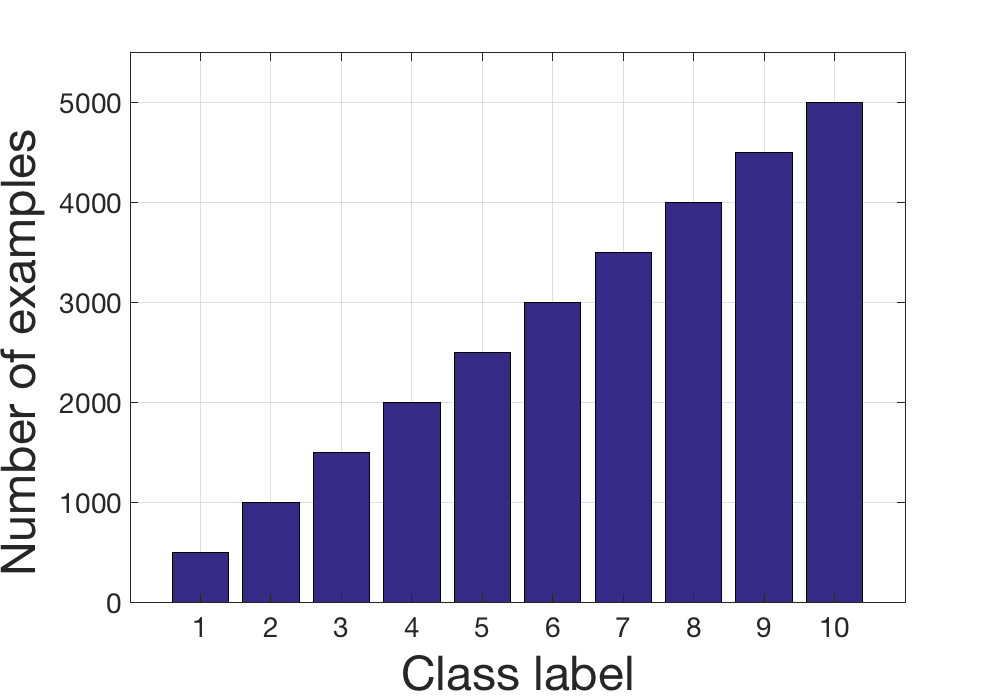}
            \caption{${\rho = 10}$}
            \label{fig:imbalance-line10}
    \end{subfigure}
    \caption{Example distributions of imbalanced set together with corresponding values of parameters $\rho$ and $\mu$ for \textit{step imbalance} (a~-~b) and $\rho$ for \textit{linear imbalance} (c).}
    \label{fig:imbalance}
\end{figure}

The first type is \textit{step imbalance}.
In \textit{step imbalance}, the number of examples is equal within minority classes and equal within majority classes but differs between the majority and minority classes.
This type of imbalance is characterized by two parameters.
One is the fraction of minority classes defined by
\begin{equation}
    \mu = \frac{|\{i \in \{1, \ldots, N\}: C_i \text{ is minority}\}|}{N},
    \label{eq:minority}
\end{equation}
where $C_i$ is a set of examples in class $i$ and $N$ is the total number of classes.
The other parameter is a ratio between the number of examples in majority classes and the number of examples in minority classes defined as follows.
\begin{equation}
    \rho = \frac{\max_{i}\{|C_i|\}}{\min_{i}\{|C_i|\}}
    \label{eq:ratio}
\end{equation}
An example of this type of imbalance is the situation when among the total of 10 classes, 5 of them have 500 training examples and another 5 have 5\,000.
In this case $\rho = 10$ and $\mu = 0.5$, as shown in Figure~\ref{fig:imbalance-step10}.
A dataset with the same number of examples in total that has smaller imbalance ratio, corresponding to parameter $\rho = 2$, but more classes being minority, $\mu = 0.9$, is presented in Figure~\ref{fig:imbalance-step2}.

The second type of imbalance we call \textit{linear imbalance}.
We define it with one parameter that is a ratio between the maximum and minimum number of examples among all classes, as in Equation~\ref{eq:ratio} for imbalance ratio in \textit{step imbalance}.
However, the number of examples in the remaining classes is interpolated linearly such that the difference between consecutive pairs of classes is constant.
An example of linear imbalance distribution with $\rho = 10$ is shown in Figure~\ref{fig:imbalance-line10}.

\subsection{Methods of addressing imbalance compared in this study}
In total, we examine seven methods to handle CNN training on a dataset with class imbalance which cover most of the commonly used approaches in the context of deep learning:
\begin{enumerate}
\itemsep0em 
    \item Random minority oversampling
    \item Random majority undersampling
    \item Two-phase training with pre-training on randomly oversampled dataset
    \item Two-phase training with pre-training on randomly undersampled dataset
    \item Thresholding with prior class probabilities
    \item Oversampling with thresholding
    \item Undersampling with thresholding
\end{enumerate}

We examine two variants of two-phase training method.
One on oversampled and the other on undersampled dataset.
For the second phase, we keep the same hyperparameters and learning rate decay policy as in the first phase.
Only the base learning rate from the first phase is multiplied by the factor of $10^{-1}$.
Regarding thresholding, this method originally uses the imbalanced training set to train a neural network.
We, in addition, combine it with oversampling and undersampling.

Selected methods are representative of the available approaches.
Sampling can be used to explicitly incorporate cost of the examples by their appearance.
It makes them one of many implementations of cost-sensitive learning~\cite{zhou2006training}.
Thresholding is another way of applying cost-sensitiveness by moving the output threshold such that higher cost examples are harder to misclassify.
Ensemble methods require training of multiple classifiers.
Because of considerable time needed to train deep models, it is often not practical and may be even infeasible to train multiple deep neural networks.
One-class methods have a very limited application to datasets with extremely high imbalance.
Moreover, they are applied to anomaly detection problem that is beyond the scope of our study.

Importantly, we focused on methods that are widely used and relatively straightforward to implement as our aim is to draw conclusions that will be practical and serve as a guidance to a large number of deep learning researchers and engineers.

\subsection{Datasets and models}
\label{sec:data}
In our study, we used three benchmark datasets: MNIST~\cite{lecun1998gradient}, CIFAR\nobreakdash-10~\cite{krizhevsky2009learning} and ImageNet Large Scale Visual Recognition Challenge (ILSVRC) 2012~\cite{russakovsky2015imagenet}.
All of them are provided with a split on training and test set that are both labeled.
For each dataset we choose different model with a set of hyperparameters used for its training that is known to perform well based on the literature.
Datasets together with their corresponding models are of increasing complexity.
This allows us to draw some conclusions on simple task and then verify how they scale to more complex ones.

All networks for the same dataset were trained with equal number of iterations.
It means that the number of epochs differs between the imbalanced versions of dataset.
This way we keep the number of weights' updates constant.
Also, all networks were trained from a random initialization of weights and no pretraining was applied.
An overview of some information about the datasets and their corresponding models is given in Table~\ref{tab:datasets}.
All experiments were implemented in the deep learning framework \textit{Caffe}~\cite{jia2014caffe}.

\begin{table}[!ht]
    \centering
    \begin{tabular}{c c c c c c c c} \toprule
     & \multicolumn{3}{c}{Image dimensions} &  & \multicolumn{2}{c}{Images per class} &  \\ \cmidrule{2-4} \cmidrule{6-7}
    Dataset & Width & Height & Depth & No. classes & Training & Test & CNN model \\ \midrule
    MNIST & 28 & 28 & 1 & 10 & 5\,000 & 1\,000 & LeNet-5 \\
    CIFAR\nobreakdash-10 & 32 & 32 & 3 & 10 & 5\,000 & 1\,000 & All-CNN \\
    ILSVRC-2012 & $\geq 256$ & $\geq 256$ & 3 & 1\,000 & 1\,000 & 50 & ResNet-10 \\ \bottomrule
    \end{tabular}
    \caption{Summary of the used datasets. The number of images per class refers to the perfectly balanced subsets used for experiments. Provided image dimensions for ImageNet are given after rescaling.}
    \label{tab:datasets}
\end{table}

\subsubsection{MNIST}
MNIST is considered simple and solved problem that involves digits’ images classification.
The dataset consists of grayscale images of size $28\times28$.
There are ten classes corresponding to digits from 0 to 9.
The number of examples per class in the original training dataset ranges from 5421 in class 5 to 6742 in class 1.
In artificially imbalanced versions we uniformly at random subsample each class to contain no more than 5\,000 examples.

The CNN model that we use for MNIST is the modern version of LeNet\nobreakdash-5~\cite{lecun1998gradient}.
The network architecture is presented in Table~\ref{tab:lenet}.
All networks for this dataset were trained for 10\,000 iterations.
Optimization algorithm is stochastic gradient descent (SGD) with momentum value of $\mu = 0.9$~\cite{qian1999momentum}.
The learning rate decay policy is defined as ${\eta_t = \eta_0 \cdot \left( 1 + \gamma \cdot t \right)^{-\alpha}}$, where ${\eta_0 = 0.01}$ is a base learning rate, ${\gamma = 0.0001}$ and ${\alpha = 0.75}$ are decay parameters and $t$ is the current iteration.
Furthermore, we used a batch size of 64 and a weight decay value of $\lambda = 0.0005$.
Network weights were initialized randomly with uniform distribution and Xavier variance~\cite{glorot2010understanding} whereas the biases were initialized with zero.
No data augmentation was used.
Test error of the model trained as described above on the original MNIST dataset was below 1\%.

\begin{table}[!ht]
    \centering
    \begin{tabular}{c c c c c c} \toprule
     & \multicolumn{3}{c}{Data dimensions} & & \\ \cmidrule{2-4}
    Layer & Width & Height & Depth & Kernel size & Stride \\ \midrule
    Input & 28 & 28 & 1 & - & - \\
    Convolution & 24 & 24 & 20 & 5 & 1 \\
    Max Pooling & 12 & 12 & 20 & 2 & 2 \\
    Convolution & 8 & 8 & 50 & 5 & 1 \\
    Max Pooling  & 4 & 4 & 50 & 2 & 2 \\
    Fully Connected & 1 & 1 & 500 & - & - \\
    ReLU & 1 & 1 & 500 & - & - \\
    Fully Connected & 1 & 1 & 10 & - & - \\
    Softmax & 1 & 1 & 10 & - & - \\ \bottomrule
    \end{tabular}
    \caption{Architecture of LeNet-5 CNN used in MNIST experiments.}
    \label{tab:lenet}
\end{table}

Experiments on MNIST dataset are performed on the following imbalance parameters space.
For \textit{linear imbalance} we test values of ${\rho \in \{ 10, 25, 50, 100, 250, 500, 1\,000, 2\,500, 5\,000 \}}$.
For \textit{step imbalance} the set of $\rho$ values is the same and for each we use all possible number of minority classes from 1 to 9, which corresponds to ${\mu \in \{ 0.1, 0.2, 0.3, 0.4, 0.5, 0.6, 0.7, 0.8, 0.9 \}}$.
The experiment for each combination of parameters is repeated 50 times.
Every time the subset of minority classes is randomized.
This way, we have created 4\,050 artificially imbalanced training sets for \textit{step imbalance} and 450 for linear imbalance.
As we have evaluated four methods that require training a model, the total number of trained networks, including baseline, is 22\,500.

\subsubsection{CIFAR-10}
CIFAR\nobreakdash-10 is a significantly more complex image classification problem than MNIST.
It contains $32\times32$ color images with ten classes of natural objects.
It does not have any natural imbalance at all.
There are exactly 5\,000 training and 1\,000 test examples in each class.
We do not use any data augmentation but follow standard preprocessing comprising global contrast normalization and ZCA whitening~\cite{goodfellow2013maxout}.

For CIFAR\nobreakdash-10 experiments we use one of the best performing type of CNN model on this dataset, i.e. All-CNN~\cite{springenberg2014striving}.
The network architecture is presented in Table~\ref{tab:allcnn}.
The networks were trained for 70\,000 iterations using SGD with momentum $\mu = 0.9$.
The base learning rate was multiplied by a fixed multiplier of 0.1 after 40\,000, 50\,000 and 60\,000 iterations.
The number of examples in a batch was 256 and a weight decay value was $\lambda = 0.001$.
Network weights were initialized with Xavier procedure and the biases set to zero.
Test error of the model trained as described above on the original CIFAR\nobreakdash-10 dataset was 9.75\%.

We have found the network training to be quite sensitive to initialization and the choice of base learning rate.
Sometimes the network gets stuck in a very poor local minimum.
Also, for more imbalanced datasets the training required lower base learning rate to train at all.
Therefore, for each case we were searching for the best one from the fixed set ${\eta_0 \in \{ 0.05, 0.005, 0.0005, 0.00005 \}}$.
Similar procedure was used by the authors of the model architecture~\cite{springenberg2014striving}.
Moreover, each training was repeated twice on the same dataset.
For a particular method and imbalanced dataset, we pick the model with the best score on the test set over all eight runs.

\begin{table}[!ht]
    \centering
    \begin{tabular}{c c c c c c c} \toprule
     & \multicolumn{3}{c}{Data dimensions} & & & \\ \cmidrule{2-4}
    Layer & Width & Height & Depth & Kernel size & Stride & Padding \\ \midrule
    Input & 32 & 32 & 3 & - & - & - \\
    Dropout (0.2) & 32 & 32 & 3 & - & - & - \\
    $2\times$(Convolution + ReLU) & 32 & 32 & 96 & 3 & 1 & 1 \\
    Convolution + ReLU & 16 & 16 & 96 & 3 & 2 & 1 \\
    Dropout (0.5) & 16 & 16 & 96 & - & - & - \\
    $2\times$(Convolution + ReLU) & 16 & 16 & 192 & 3 & 1 & 1 \\
    Convolution + ReLU & 8 & 8 & 192 & 3 & 2 & 1 \\
    Dropout (0.5) & 8 & 8 & 192 & - & - & - \\
    Convolution + ReLU & 6 & 6 & 192 & 3 & 1 & 0 \\
    Convolution + ReLU & 6 & 6 & 192 & 1 & 1 & 0 \\
    Convolution + ReLU & 6 & 6 & 10 & 1 & 1 & 0 \\
    Average Pooling & 1 & 1 & 10 & 6 & - & - \\
    Softmax & 1 & 1 & 10 & - & - & -\\ \bottomrule
    \end{tabular}
    \caption{Architecture of All-CNN used in CIFAR\nobreakdash-10 experiments.}
    \label{tab:allcnn}
\end{table}

The network architecture does not have any fully connected layers.
Therefore, during the fine-tuning in two-phase training method we update the weights of two last convolutional layers with kernels of size 1.

The imbalance parameters space used in CIFAR\nobreakdash-10 experiments is considerably sparser than the one used for MNIST due to the significantly longer time required to train one network.
The set of tested values was narrowed to make the experiment run in a reasonable time.
For \textit{linear} and \textit{step imbalance}, we test values of ${\rho \in \{ 2, 10, 20, 50 \}}$.
In \textit{step imbalance}, for each value of $\rho$, the set of values of parameter $\mu$ was ${\mu \in \{ 0.2, 0.5, 0.8 \}}$, which corresponds to having two, five and eight minority classes, respectively.
And for all the cases, the classes chosen to be minority were the ones with the lowest label value.
It means that for a fixed number of minority classes the same classes were always picked as minority.
Also, all of them were included in a larger set of minority classes.
In total we trained 640 networks on this dataset.

\subsubsection{ImageNet}
For evaluation we use a ILSVRC\nobreakdash-2012 competition subset of ImageNet, widely used as a benchmark to compare classifiers' performance.
The number of examples in majority classes was reduced from 1\,200 to 1\,000.
Classes with less than 1\,000 cases were always chosen as a minority ones for imbalanced subsets.
The only data preprocessing applied is resizing such that the smaller dimension is 256 pixels long and the aspect ratio is preserved.
During training, as input we use a randomly cropped $224\times224$ pixel square patch and a single centered crop in a test phase.
Moreover, during training we randomly mirror images, but there is no color, scale or aspect ratio augmentation.

\begin{figure}[!ht]
    \centering
    \includegraphics[width=\linewidth]{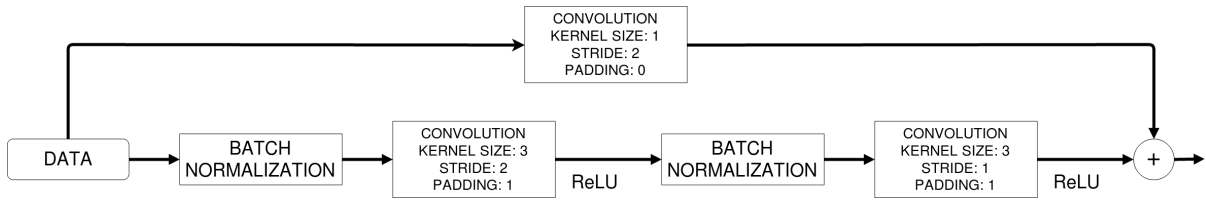}
    \caption{Architecture of a single residual block in ResNet used in ILSVRC-2012 experiments.}
    \label{fig:resnet}
\end{figure}

A model architecture employed for this dataset is ResNet\nobreakdash-10~\cite{simon2016imagenet}, i.e. a residual network~\cite{he2016deep} with batch normalization layers that are known to accelerate deep networks training~\cite{ioffe2015batch}.
It consists of four residual blocks that give us nine convolutional layers and one fully connected.
The first residual block outputs data tensor of depth 64 and then each one increases it by a factor of two.
Fully connected layer outputs 1\,000 values to softmax that transforms them to class probabilities.
The architecture of one residual block is presented in Figure~\ref{fig:resnet}.

The networks were trained for 320\,000 iterations using SGD with momentum $\mu = 0.9$.
The base learning rate is set to ${\eta_0 = 0.1}$ and decays linearly to 0 in the last iteration.
The number of examples in a batch was $256$ and a weight decay $\lambda = 0.0001$.
Network weights were initialized with Kaiming (also known as MSRA) initialization procedure~\cite{he2015delving}.
Top-1 test error of the model trained as described above on the original ILSVRC\nobreakdash-2012 dataset was 62.56\% and 99.50 multi-class ROC AUC for a single centered crop.

We have chosen relatively small ResNet for the sake of faster training~\footnote{It takes five days to train one ResNet\nobreakdash-10 network on Nvidia GTX 1070 GPU.}.
We test only one case of small and two cases of large \textit{step imbalance} and run it on the baseline, undersampling and oversampling methods.
Specifically, all three \textit{step imbalanced} subsets are defined with ${\mu = 0.1, \rho = 10}$, ${\mu = 0.8, \rho = 50}$ and ${\mu = 0.9, \rho = 100}$.
They correspond to 100 minority classes with imbalance ratio of 10, 800 minority classes with imbalance of 50, and 900 minority classes with imbalance ratio of 100, respectively.
Moreover, for the highest imbalance, we train three networks for each method with randomized selection of minority classes and subsampled set of examples in each class.
This is done in order to estimate variability in performance of methods.
In total, this gives us 15 ResNet\nobreakdash-10 networks trained on five artificially imbalanced subsets of ILSVRC\nobreakdash-2012.

\subsection{Evaluation metrics and testing}
\label{sec:evaluation}
The metric that is most widely used to evaluate a classifier performance in the context of multiclass classification with CNNs is overall accuracy which is the proportion of test examples that were correctly classified.
However, it has some significant and long acknowledged limitations, particularly in the context of imbalanced datasets~\cite{chawla2005data}.
Specifically, when the test set is imbalanced, accuracy will favor classes that are overrepresented in some cases leading to highly misleading assessment.
An example of this is a situation when the majority class represents 99\% of all cases and the classifier assigns the label of the majority class to all test cases.
A misleading accuracy of 99\% will be assigned to a classifier that has a very limited use.
Another issue might arise when the test set is balanced and a training set is imbalanced.
This might result in a situation when a decision threshold is moved to reflect the estimated class prior probabilities and cause a low accuracy measure in the test set while the true discriminative power of the classifier does not change.

A measure that addresses these issues is area under the receiver operating characteristic curve (ROC AUC)~\cite{bradley1997use} which is a plot of the false positive rate to the true positive rate for all possible prediction thresholds.
We used a specific implementation of the ROC AUC available in scikit-learn python package~\cite{pedregosa2011scikit}.
It calculates sensitivities and specificities at all thresholds defined by the responses of the classifier in the test set followed by the AUC calculation using the trapezoid rule.
ROC AUC is a well-studied and sound measure of discrimination~\cite{ling2003auc} and has been widely used as an evaluation metric for classifiers.
ROC has also been used to compare performance of classifiers trained on imbalanced datasets~\cite{mazurowski2008training, maloof2003learning}.
Since the basic version of ROC is only suitable for binary classification, we use a multi-class modification of it~\cite{provost2003tree}.
The multi-class ROC is calculated by taking the average of AUCs obtained independently for each class for the binary classification task of distinguishing a given class from all the other classes.

Test set of all used datasets has equal number of examples in each class.
Usually, it is assumed that the class distribution of a test set follows the one of a training set.
We do not change a test set to match artificially imbalanced training set.
The reason is that the score achieved by each classifier on the same test set is more comparable and the largest number of cases in each of the classes provides the most accurate performance estimation.

\begin{figure}[!ht]
    \centering
    \begin{subfigure}[b]{0.9\textwidth}
            \includegraphics[width=\linewidth]{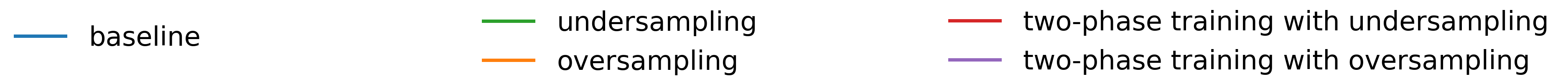}
    \end{subfigure}
    \vspace{0.5em}
    
    \begin{subfigure}[b]{0.32\textwidth}
            \includegraphics[width=\linewidth]{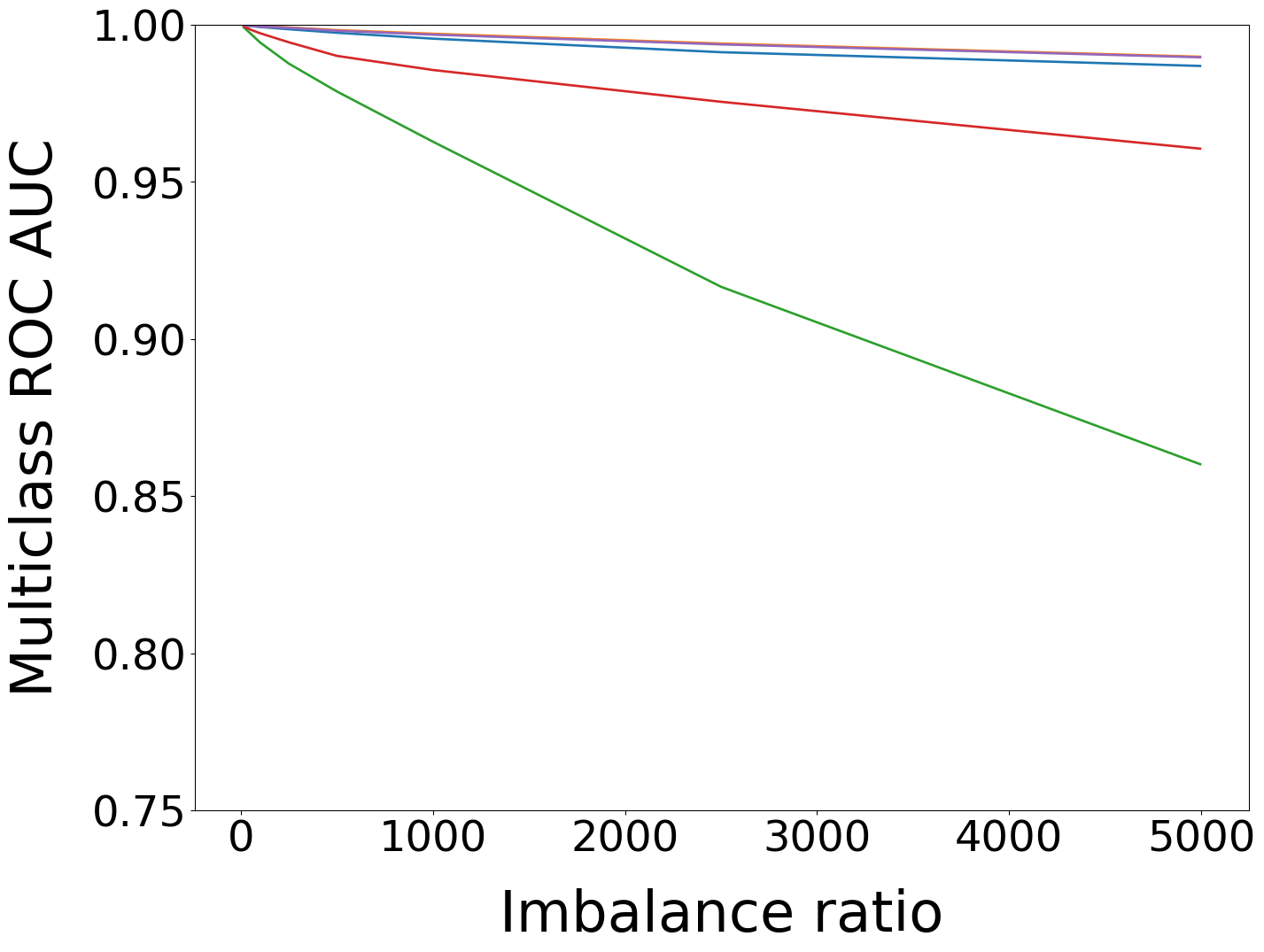}
            \caption{2 minority classes}
            \label{fig:mnist-step_multi_roc_auc_2min}
    \end{subfigure}
    \begin{subfigure}[b]{0.32\textwidth}
            \includegraphics[width=\linewidth]{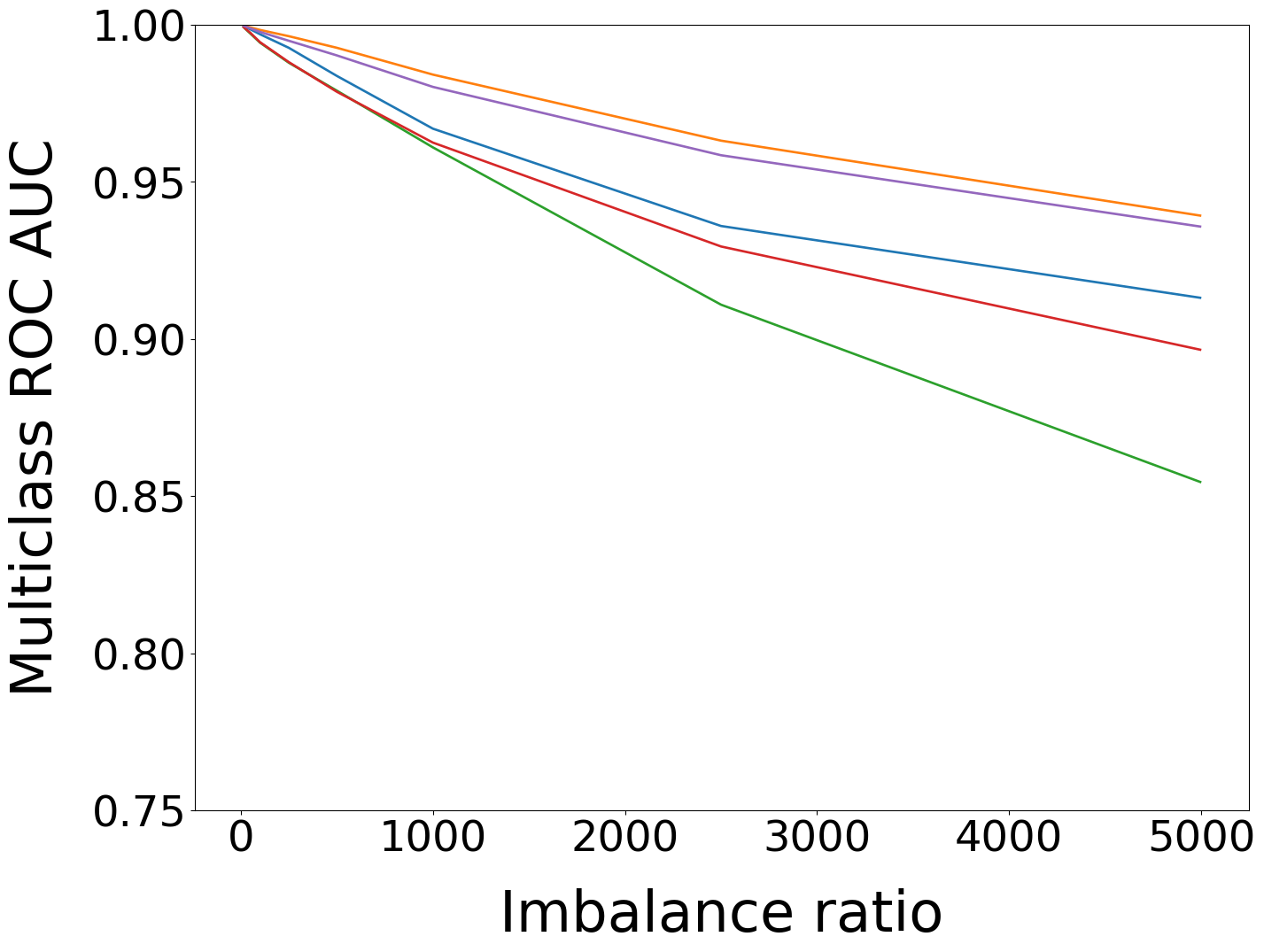}
            \caption{5 minority classes}
            \label{fig:mnist-step_multi_roc_auc_5min}
    \end{subfigure}
    \begin{subfigure}[b]{0.32\textwidth}
            \includegraphics[width=\linewidth]{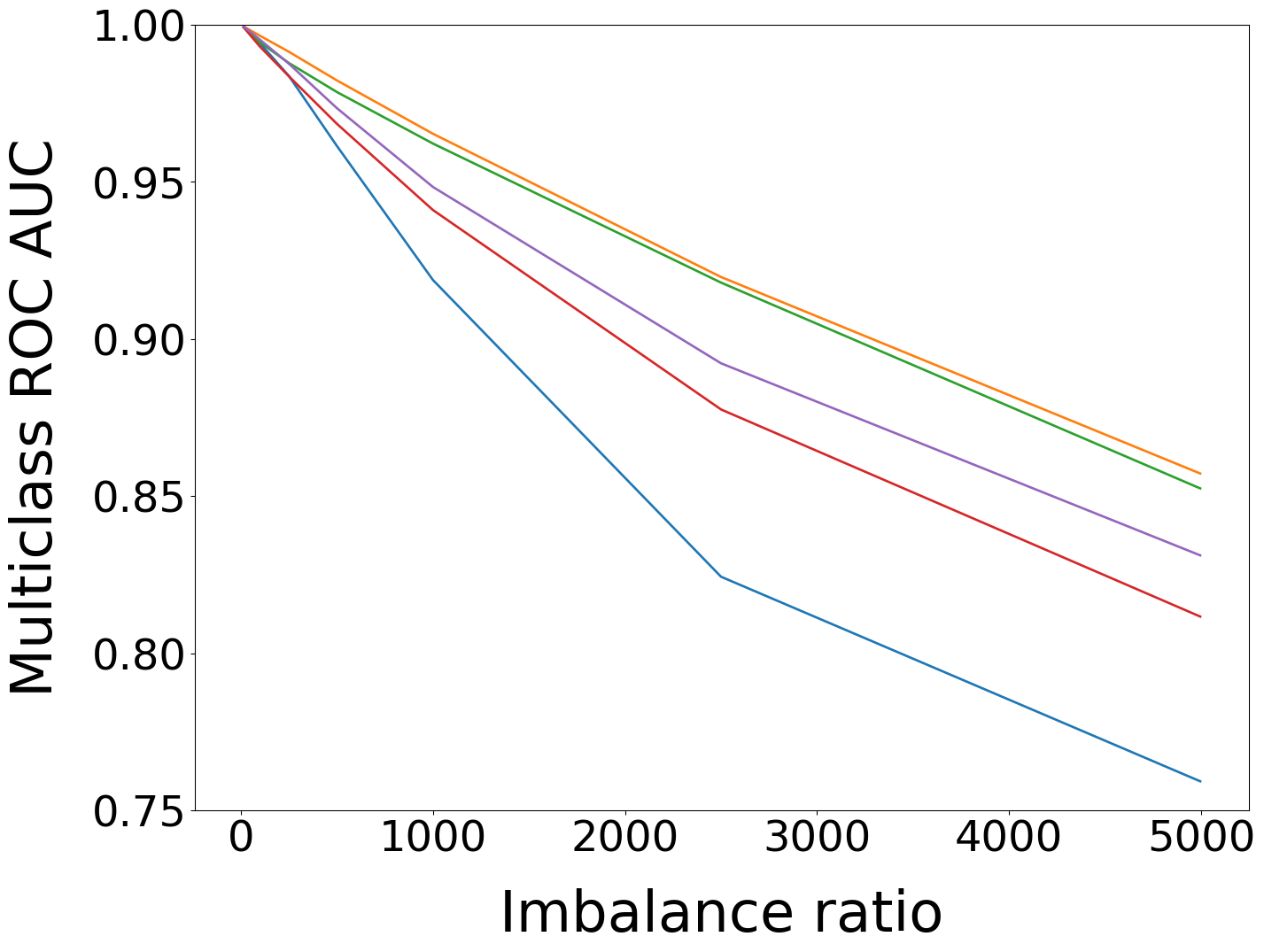}
            \caption{8 minority classes}
            \label{fig:mnist-step_multi_roc_auc_8min}
    \end{subfigure}
    \vspace{0.5em}
    
    \begin{subfigure}[b]{0.32\textwidth}
            \includegraphics[width=\linewidth]{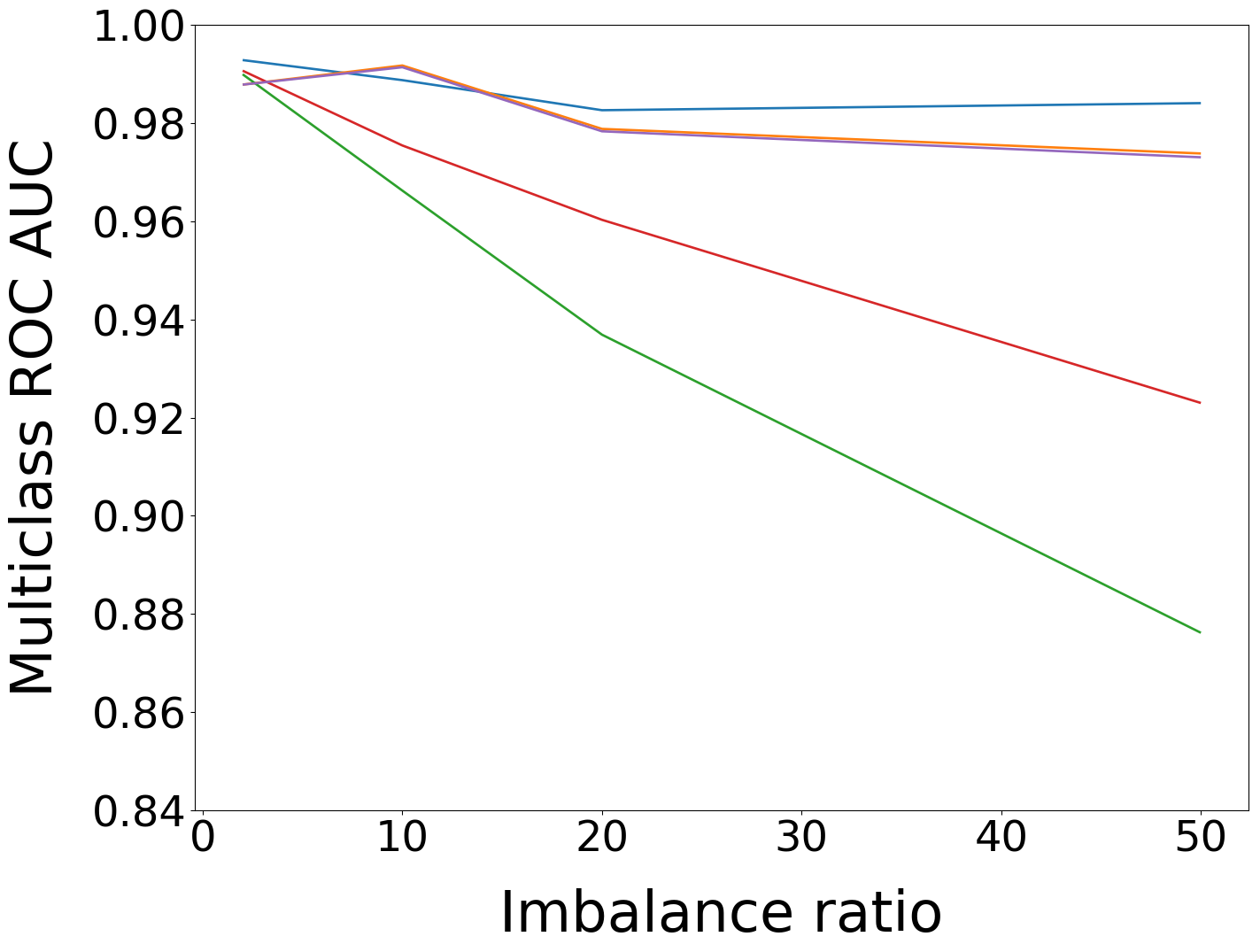}
            \caption{2 minority classes}
            \label{fig:cifar-step_multi_roc_auc_2min}
    \end{subfigure}
    \begin{subfigure}[b]{0.32\textwidth}
            \includegraphics[width=\linewidth]{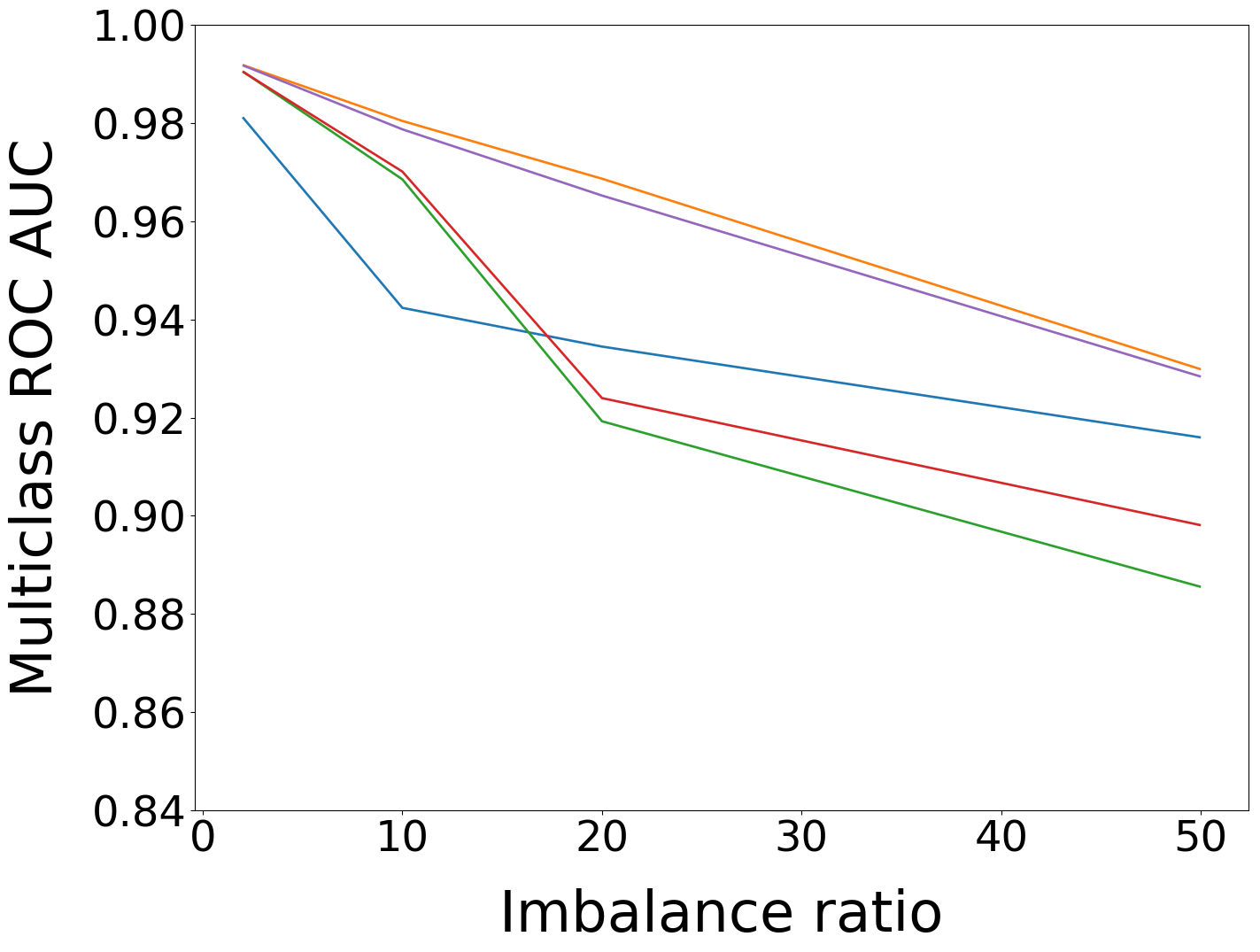}
            \caption{5 minority classes}
            \label{fig:cifar-step_multi_roc_auc_5min}
    \end{subfigure}
    \begin{subfigure}[b]{0.32\textwidth}
            \includegraphics[width=\linewidth]{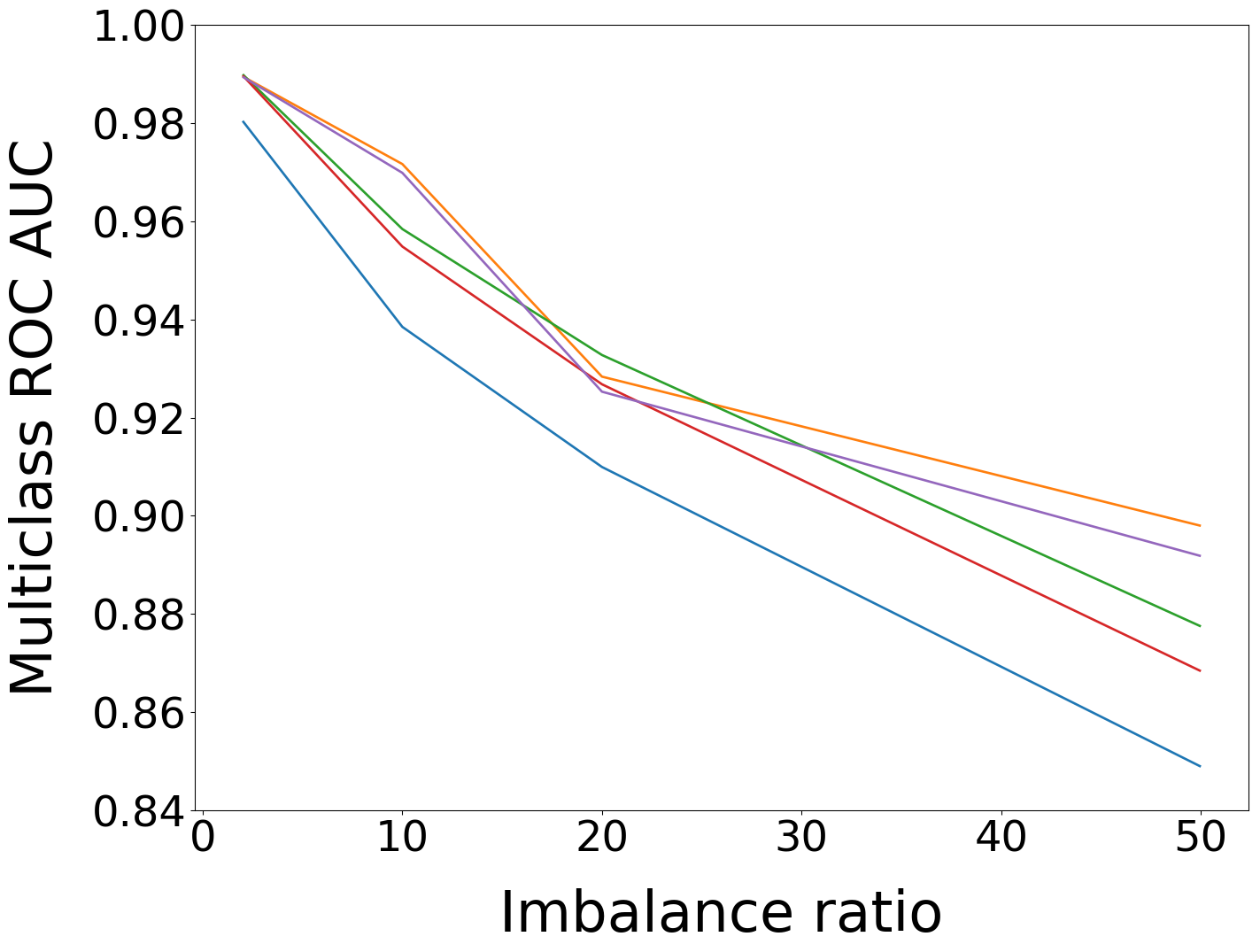}
            \caption{8 minority classes}
            \label{fig:cifar-step_multi_roc_auc_8min}
    \end{subfigure}
    \caption{Comparison of methods with respect to multi-class ROC AUC on MNIST (a~-~c) and CIFAR\nobreakdash-10 (d~-~f) for \textit{step imbalance} with fixed number of minority classes.}
    \label{fig:step_multi_roc_auc_min}
\end{figure}

\section{Results}
\label{sec:results}

\subsection{Effects of class imbalance on classification performance and comparison of methods to address imbalance}
\label{sec:effect}
\begin{figure}[!ht]
    \centering
    \begin{subfigure}[b]{0.9\textwidth}
            \includegraphics[width=\linewidth]{legend/step_multi_roc_auc_legend}
    \end{subfigure}
    \vspace{0.5em}
    
    \begin{subfigure}[b]{0.32\textwidth}
            \includegraphics[width=\linewidth]{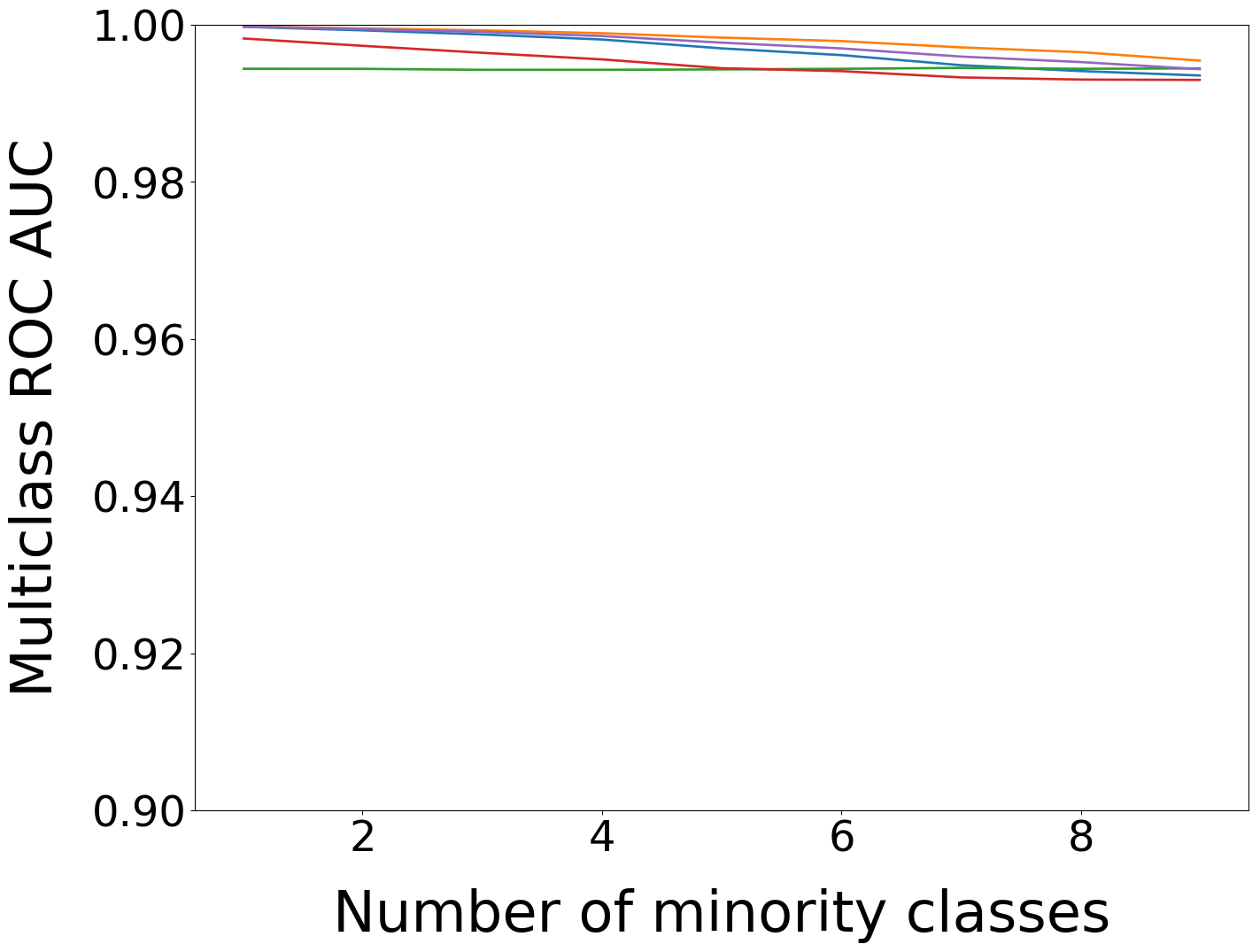}
            \caption{Imbalance ratio of 100}
            \label{fig:mnist-step_multi_roc_auc_100r}
    \end{subfigure}
    \begin{subfigure}[b]{0.32\textwidth}
            \includegraphics[width=\linewidth]{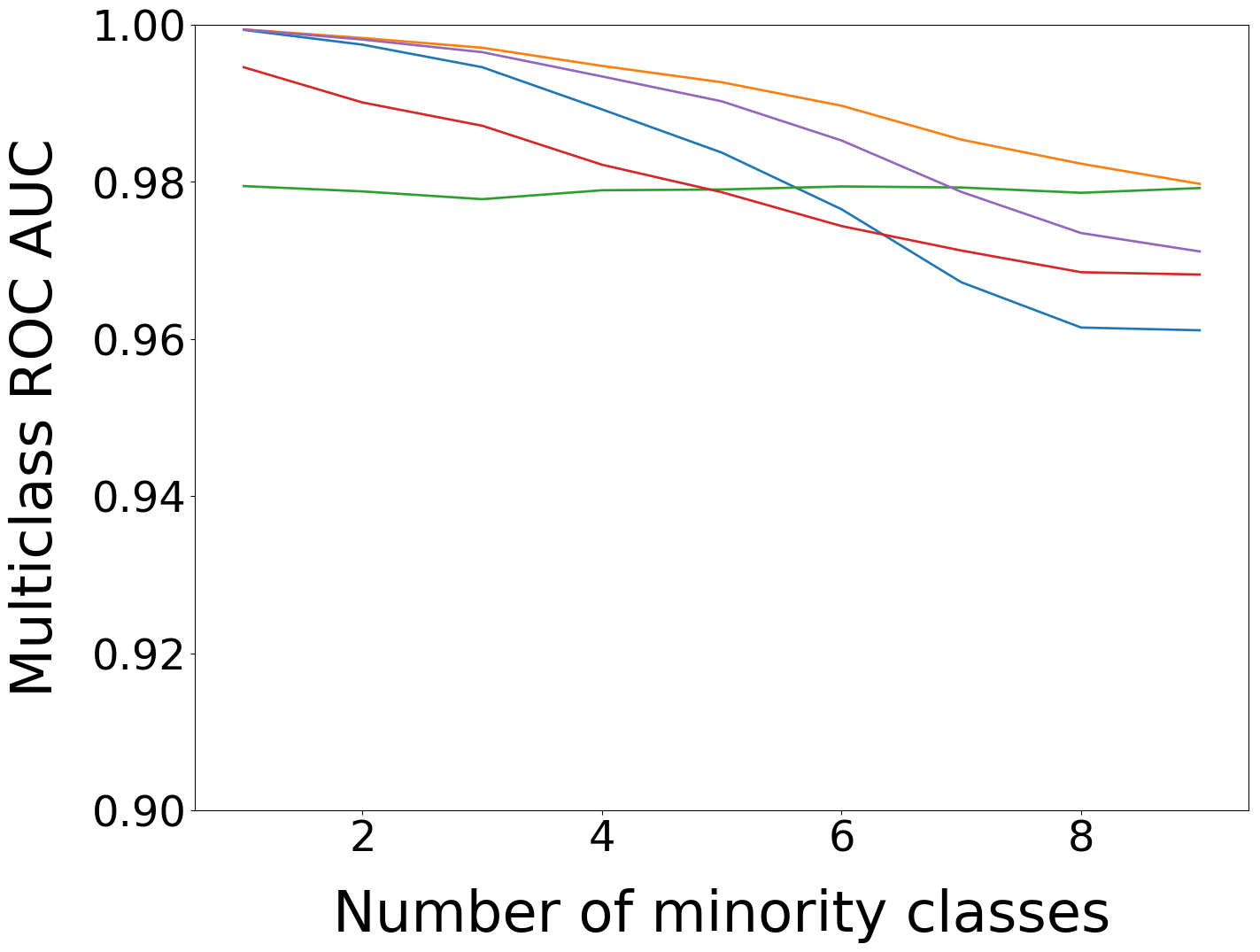}
            \caption{Imbalance ratio of 500}
            \label{fig:mnist-step_multi_roc_auc_500r}
    \end{subfigure}
    \begin{subfigure}[b]{0.32\textwidth}
            \includegraphics[width=\linewidth]{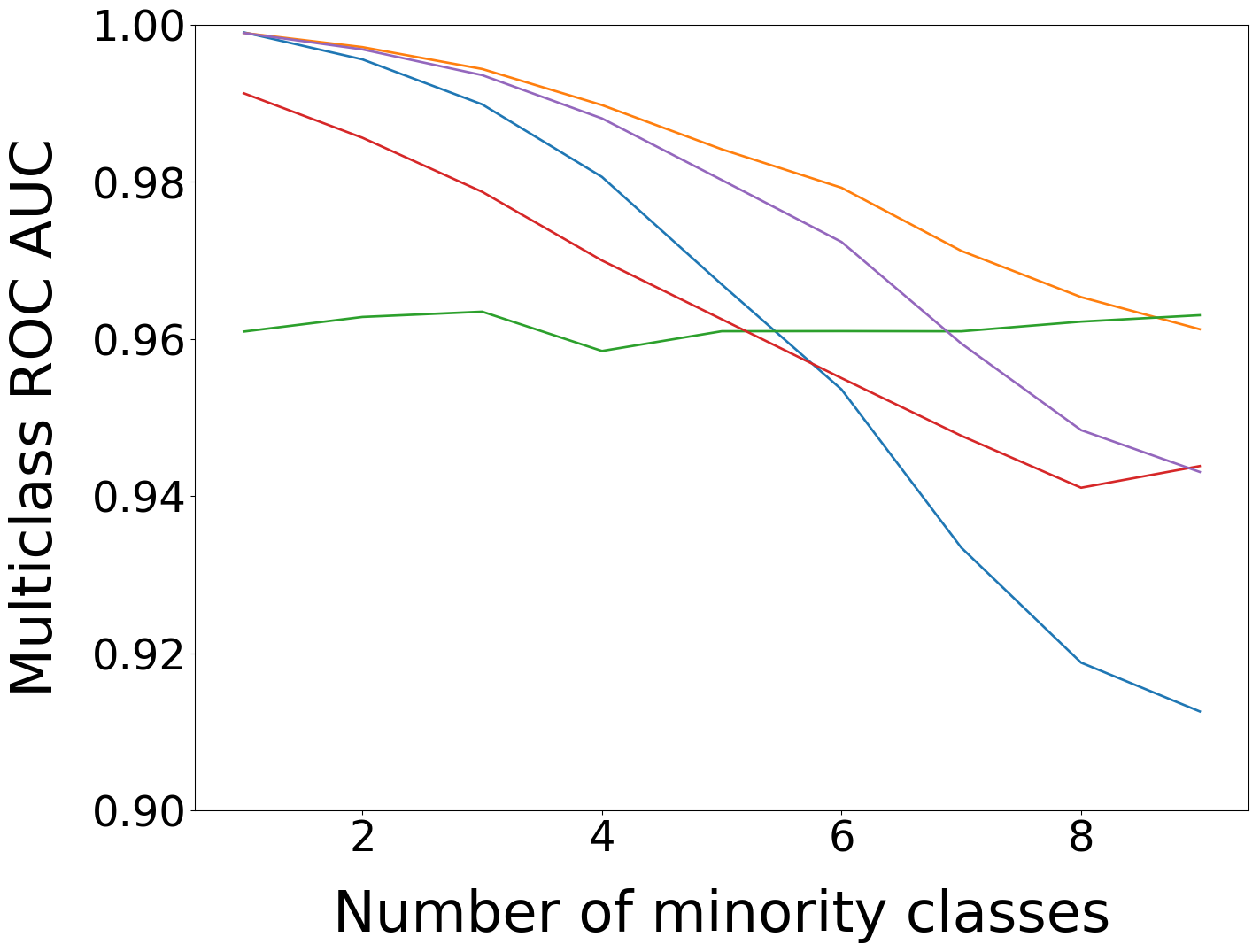}
            \caption{Imbalance ratio of 1\,000}
            \label{fig:mnist-step_multi_roc_auc_1000r}
    \end{subfigure}
    \vspace{0.5em}

    \begin{subfigure}[b]{0.32\textwidth}
            \includegraphics[width=\linewidth]{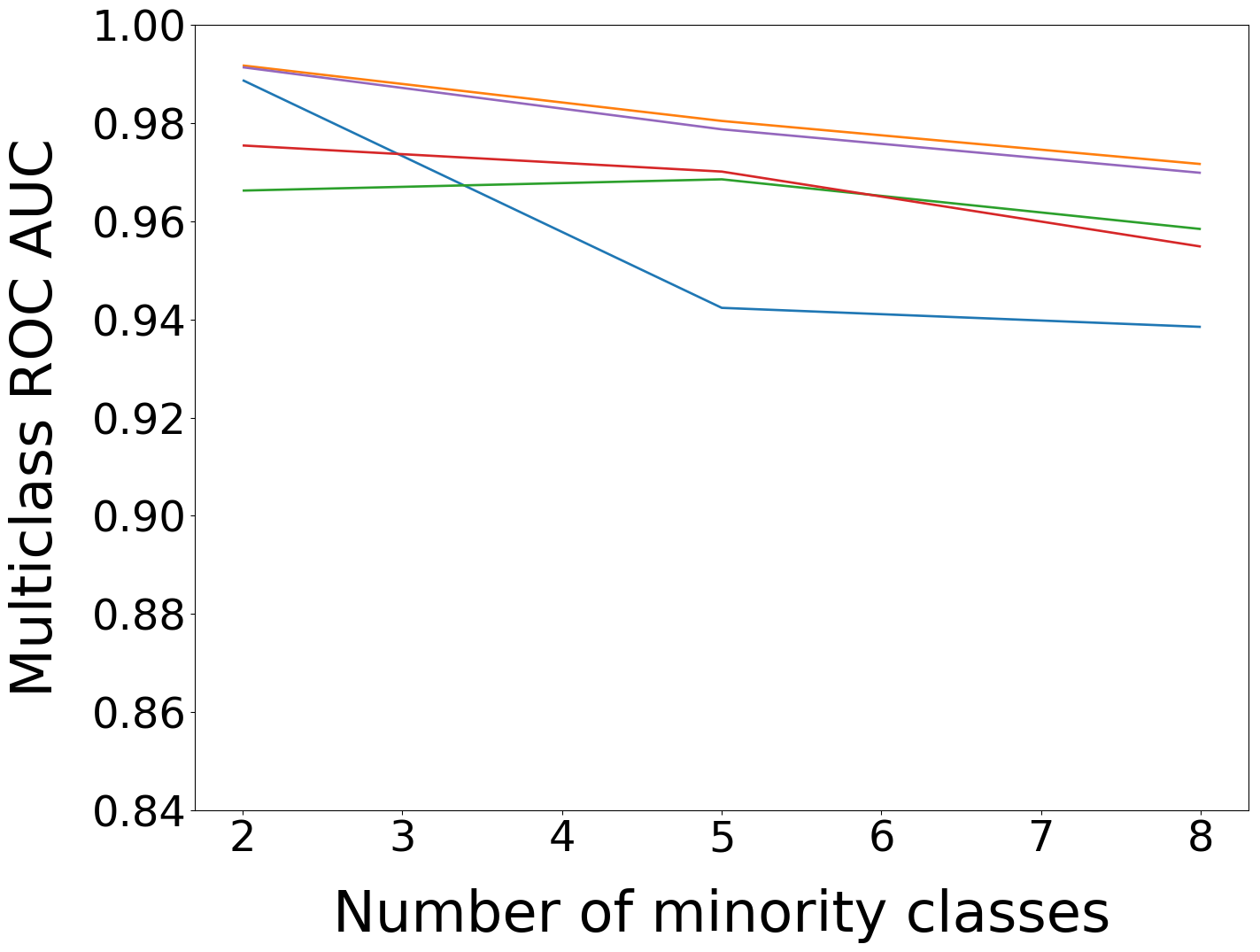}
            \caption{Imbalance ratio of 10}
            \label{fig:cifar-step_multi_roc_auc_10r}
    \end{subfigure}
    \begin{subfigure}[b]{0.32\textwidth}
            \includegraphics[width=\linewidth]{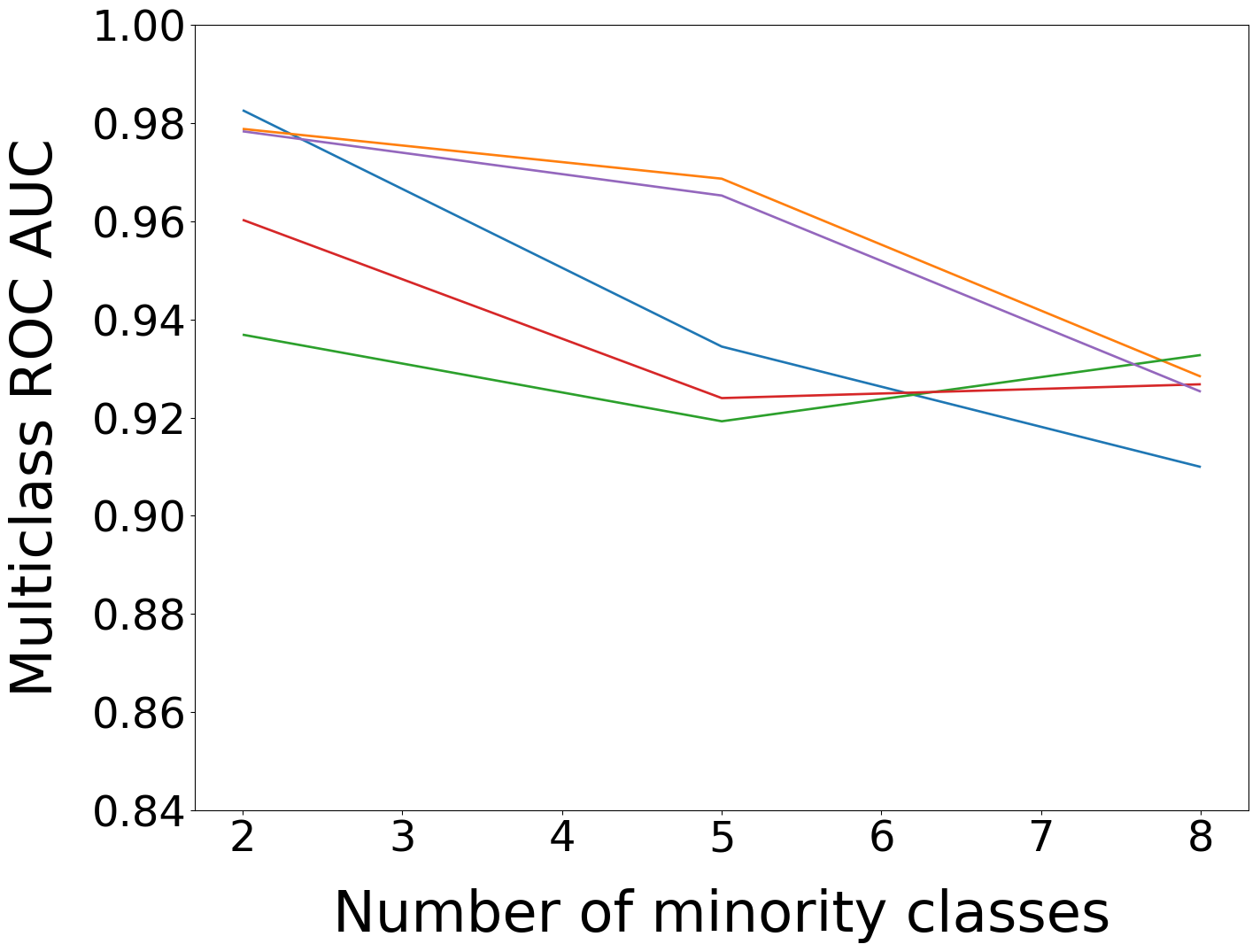}
            \caption{Imbalance ratio of 20}
            \label{fig:cifar-step_multi_roc_auc_20r}
    \end{subfigure}
    \begin{subfigure}[b]{0.32\textwidth}
            \includegraphics[width=\linewidth]{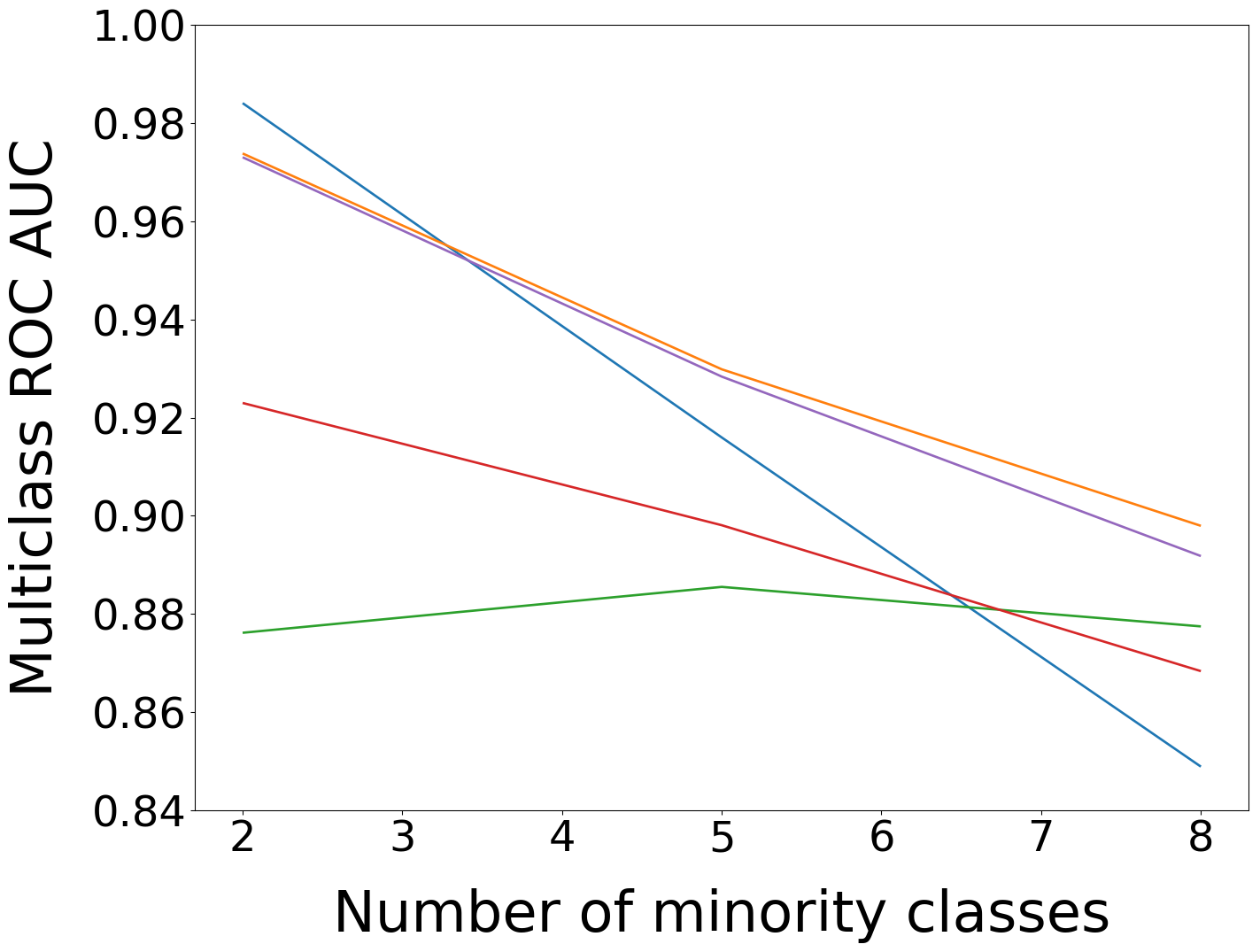}
            \caption{Imbalance ratio of 50}
            \label{fig:cifar-step_multi_roc_auc_50r}
    \end{subfigure}
    \caption{Comparison of methods with respect to multi-class ROC AUC on MNIST (a~-~c) and CIFAR\nobreakdash-10 (d~-~f) for \textit{step imbalance} with fixed imbalance ratio.}
    \label{fig:step_multi_roc_auc_ratio}
\end{figure}

The results showing the impact of class imbalance on classification performance and comparison of methods for addressing imbalance are shown in Figures~\ref{fig:step_multi_roc_auc_min} and~\ref{fig:step_multi_roc_auc_ratio}.
Figure~\ref{fig:step_multi_roc_auc_min} shows the results with respect to multi-class ROC AUC for a fixed number of minority classes on MNIST and CIFAR\nobreakdash-10.
Figure~\ref{fig:step_multi_roc_auc_ratio} presents the result from the perspective of fixed ratio of imbalance, i.e. parameter $\rho$, for the same two datasets.

Regarding the effect of class imbalance on classification performance, we observed the following.
First, the deterioration of performance due to class imbalance is substantial.
As expected, the increasing ratio of examples between majority and minority classes as well as the number of minority classes had a negative effect on performance of the resulting classifiers.
Furthermore, by comparing the results from MNIST and CIFAR\nobreakdash-10 we observed that the effect of imbalance is significantly stronger for the task with higher complexity.
A similar drop in performance for MNIST and CIFAR\nobreakdash-10 corresponded to approximately 100 times stronger level of imbalance in the MNIST dataset.

Regarding performance of different methods for addressing imbalance, in almost all of the situations oversampling emerged as the best method.
It also showed notable improvement of performance over the baseline (i.e. do-nothing strategy) in majority of the situations and never showed a considerable decrease in performance for the two datasets analyzed in this section making it a clear recommendation for tasks similar to MNIST and CIFAR\nobreakdash-10.

Undersampling showed a generally poor performance.
In a large number of analyzed scenarios, undersampling showed decrease in performance as compared to the baseline.
In scenarios with a large proportion of minority classes undersampling showed some improvement over the baseline but never a notable advantage over oversampling (Figure~\ref{fig:step_multi_roc_auc_min}).

For a fixed imbalance ratio undersampling is always trained on the subset of equal size.
As a result, its performance does not change with the number of minority classes.
For both datasets and each case of imbalance ratio, the gap between undersampling and oversampling is the biggest for smaller number of minority classes and decreases with the number of minority classes, as shown in Figure~\ref{fig:step_multi_roc_auc_ratio}.
This is expected since with all classes being minority these two methods become equivalent.

Two-phase training methods with both undersampling and oversampling tend to perform between the baseline and their corresponding method (undersampling or oversampling).
If the baseline is better than one of these methods, fine-tuning improves the original method.
Otherwise, performance deteriorates.
However, if the baseline is better, there is still no gain from using two-phase training method.
As oversampling is almost always better than the baseline, fine-tuning always gives lower score.

The variability of patters, visual structures, and objects in CIFAR\nobreakdash-10 is considerably higher then in MNIST.
For this reason, we run the \textit{step imbalance} experiment three times on a reshuffled stratified training and test split to validate our results.
Additional results are available in Appendix~A.

\begin{figure}[!ht]
    \centering
    \begin{subfigure}[b]{0.42\textwidth}
            \includegraphics[width=\linewidth]{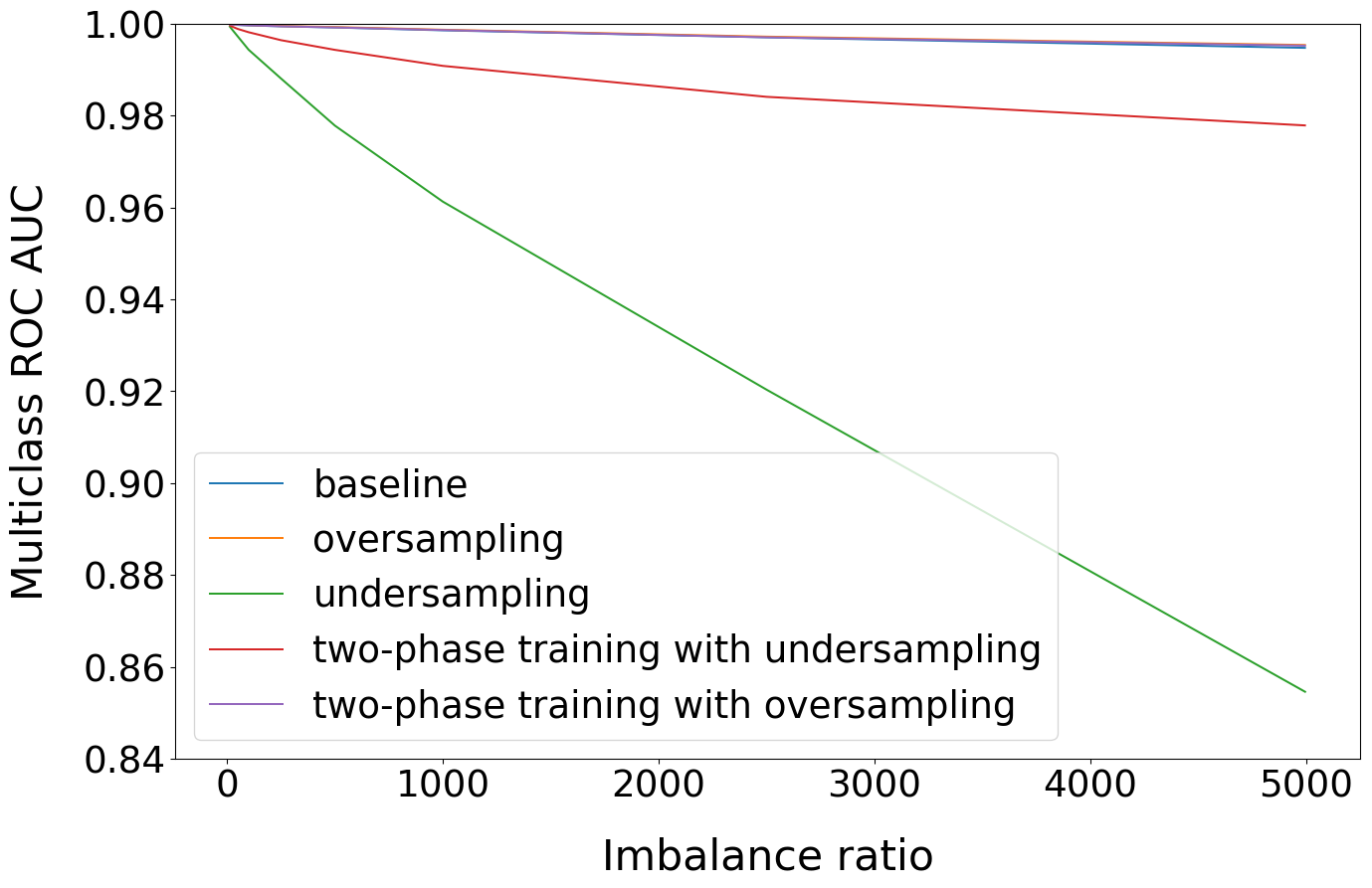}
            \caption{MNIST}
            \label{fig:mnist-lin_multi_roc_auc}
    \end{subfigure}
    \hspace{1em}
    \begin{subfigure}[b]{0.42\textwidth}
            \includegraphics[width=\linewidth]{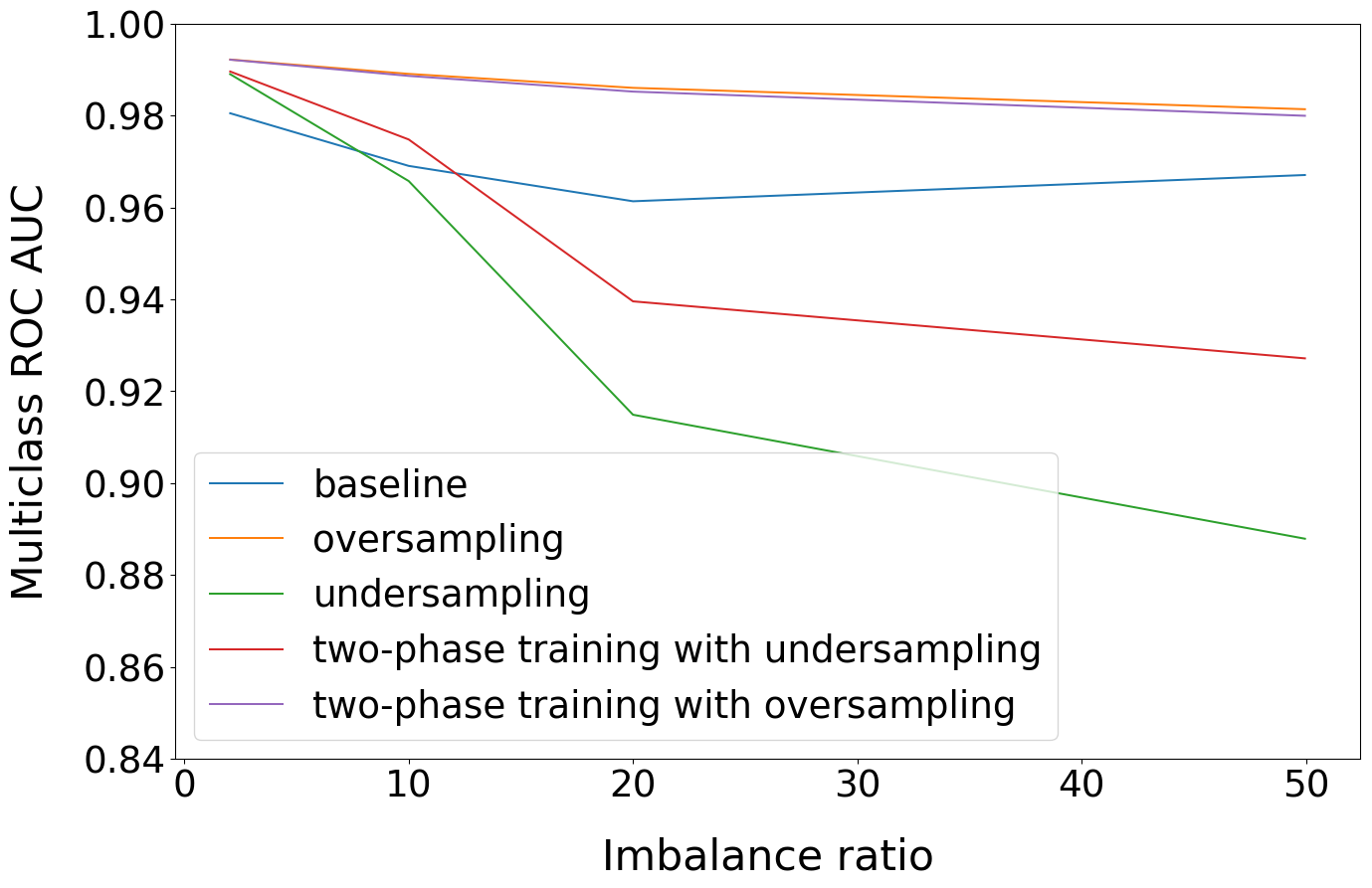}
            \caption{CIFAR-10}
            \label{fig:cifar-lin_multi_roc_auc}
    \end{subfigure}
    \caption{Comparison of methods with respect to multi-class ROC AUC for \textit{linear imbalance}.}
    \label{fig:lin_multi_roc_auc}
\end{figure}

In Figure~\ref{fig:lin_multi_roc_auc} we show the results for \textit{linear imbalance} on MNIST and CIFAR\nobreakdash-10 datasets.
The highest possible \textit{linear imbalance} ratio for MNIST dataset is 5\,000, which means only one example in the most underrepresented class.
However, even in this case the decrease in performance according to multi-class ROC AUC score for the baseline model is not significant, as shown in Figure~\ref{fig:mnist-lin_multi_roc_auc}.
Nevertheless, oversampling improves the score on both datasets and for all tested values of $\rho$, whereas the score for undersampling decreases approximately linearly with imbalance ratio.

\subsection{Results on ImageNet dataset}
\label{sec:imagenet}
The results from experiments performed on ImageNet (ILSVRC\nobreakdash-2012) dataset confirm the impact of imbalance on classifier’s performance.
Table~\ref{tab:imagenet_rocauc} compares methods with respect to multi-class ROC AUC.
The drop in performance for the largest tested imbalance was from 99 to 90, in terms of multi-class ROC AUC.
The results confirm that the oversampling approach performs consistently better than undersampling approach across all scenarios.
A small decrease in performance as compared to baseline was observed for oversampling for extreme imbalances.
Please note, however, that these results should be treated with caution and not as strong evidence that oversampling is inferior for highly complex tasks with extreme imbalance.
The absolute difference in performance between three runs with respect to multi-class ROC AUC was even higher than 4 (for undersampling).
Therefore, differences of 1~-~2 might be due to variability of results between different runs of neural networks.
Moreover, the highest tested imbalanced training set was only about 10\% of the original ILSVRC\nobreakdash-2012 introducing confounding issues such as the optimal training hyperparameters for this significantly changed dataset.
Therefore, while these results indicate that caution should be taken when any sampling technique is applied to highly complex tasks with extreme imbalances, it needs a more extensive study devoted to this specific issue.

\begin{table}[!ht]
    \centering
    \begin{tabular}{ c  c  c  r c l } \toprule
    Method & $\mu = 0.1, \rho = 10$ & $\mu = 0.8, \rho = 50$ & \multicolumn{3}{c}{$\mu = 0.9, \rho = 100$} \\ \midrule
    Baseline & 99.41 & 96.31 & 90.74 & 90.46 & 90.05 \\
    Oversampling & 99.35 & 95.06 & 88.38 & 88.39 & 88.17 \\
    Undersampling & 96.85 & 94.98 & 88.35 & 84.08 & 83.74 \\ \bottomrule
    \end{tabular}
    \caption{Comparison of results on ImageNet with respect to multi-class ROC AUC.}
    \label{tab:imagenet_rocauc}
\end{table}

\subsection{Separation of effects from reduced number of examples and class imbalance}
\label{sec:separation}
An important question that needs to be considered in the context of our study is whether the decrease in performance for imbalanced datasets is merely caused by the fact that our imbalanced datasets simply had fewer training examples or is it truly caused by the fact that the datasets are imbalanced.

First, we notice that oversampling method uses the same amount of data as the baseline.
It only eliminates the imbalance which is enough to improve the performance in almost all the cases.
Still, it does not reach the performance of a classifier trained on the original dataset.
This is an indication that the effect of imbalance is not trivial.

Second, for some cases undersampling, which reduces the total number of cases performs better than the baseline (see Figures \ref{fig:mnist-step_multi_roc_auc_8min} and \ref{fig:cifar-step_multi_roc_auc_8min}).
Moreover, there are even cases when undersampling can perform on a par with oversampling.
It means that, between two sampling methods that eliminate imbalance, even using fewer data can be comparable.

In addition, for the same value of parameter $\rho$ we have equal number of examples in the training set for \textit{linear imbalance} and \textit{step imbalance} with $\mu = 0.5$, which corresponds to half of the classes being minority.
The drop in performance is much higher for \textit{step imbalance}.
This additionally demonstrates that not only the total number of examples matters but also its distribution between classes.

\subsection{Improving accuracy score with multi-class thresholding}
\label{sec:improving}
\begin{figure}[!ht]
    \centering
    \vspace{-0.5em}
    \begin{subfigure}[b]{0.9\textwidth}
            \includegraphics[width=\linewidth]{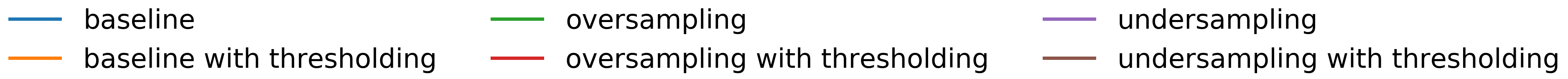}
    \end{subfigure}
    \vspace{0.5em}
    
    \begin{subfigure}[b]{0.32\textwidth}
            \includegraphics[width=\linewidth]{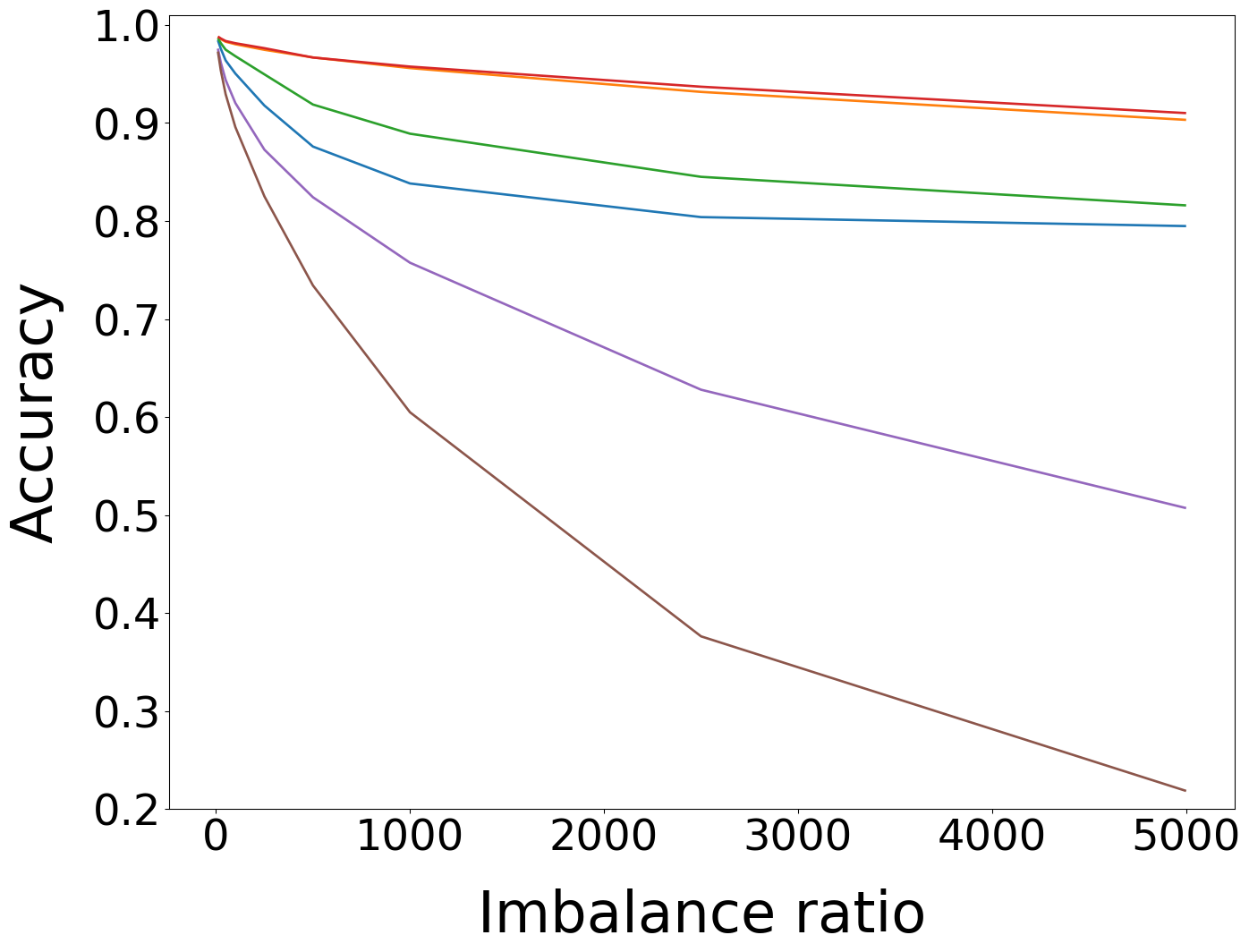}
            \caption{2 minority classes}
            \label{fig:mnist-step_acc_2min}
    \end{subfigure}
    \begin{subfigure}[b]{0.32\textwidth}
            \includegraphics[width=\linewidth]{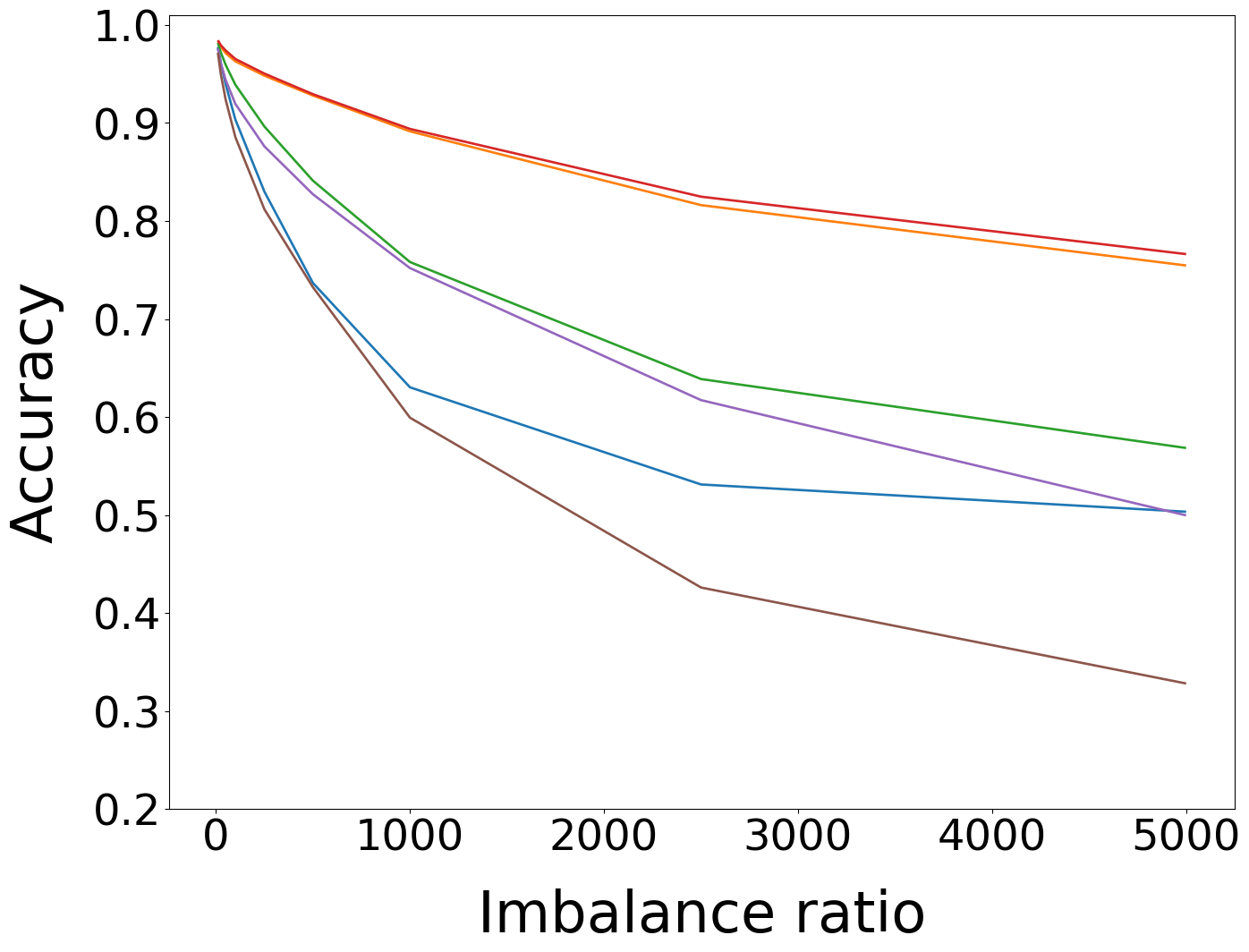}
            \caption{5 minority classes}
            \label{fig:mnist-step_acc_5min}
    \end{subfigure}
    \begin{subfigure}[b]{0.32\textwidth}
            \includegraphics[width=\linewidth]{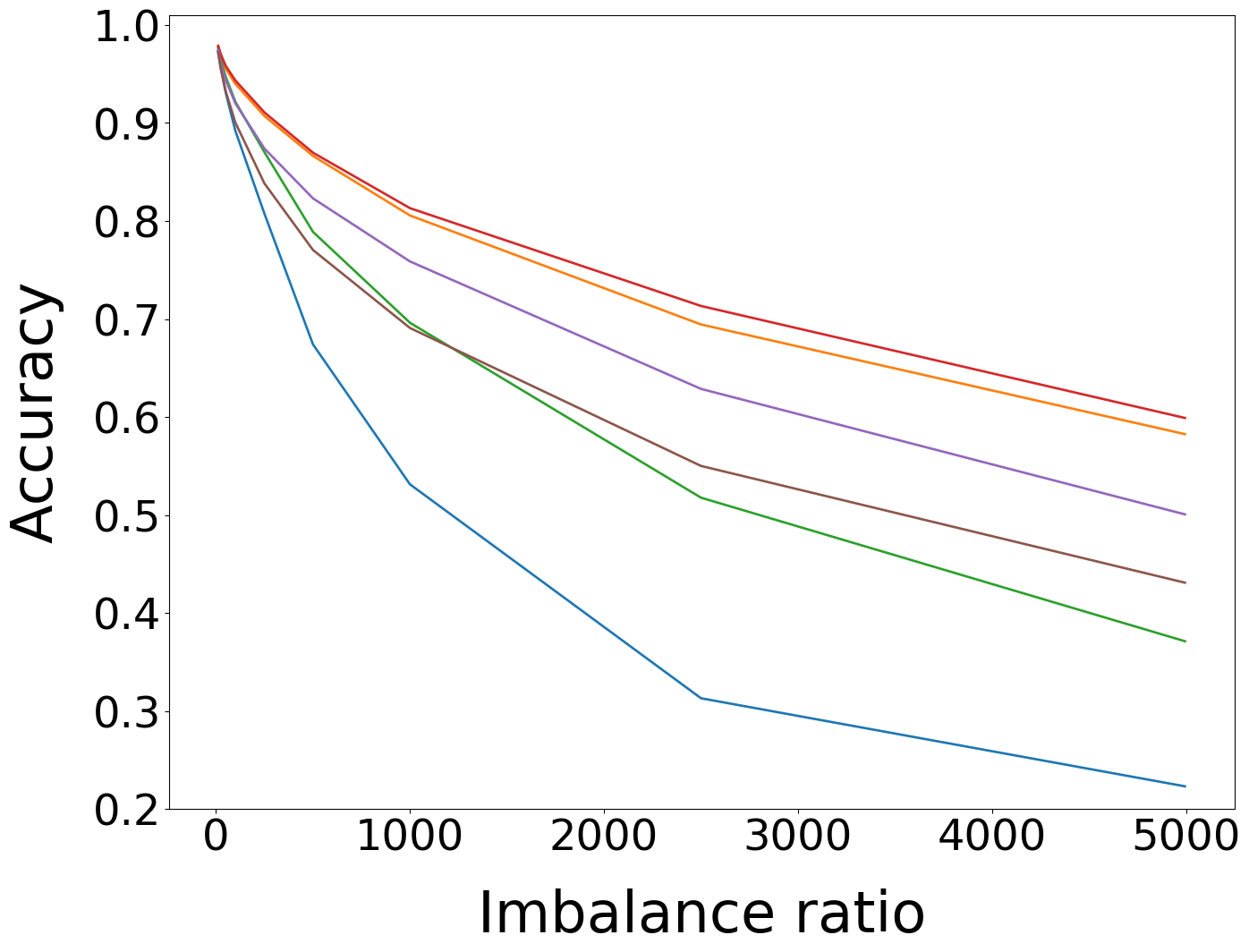}
            \caption{8 minority classes}
            \label{fig:mnist-step_acc_8min}
    \end{subfigure}
    \vspace{0.5em}
    
    \begin{subfigure}[b]{0.32\textwidth}
            \includegraphics[width=\linewidth]{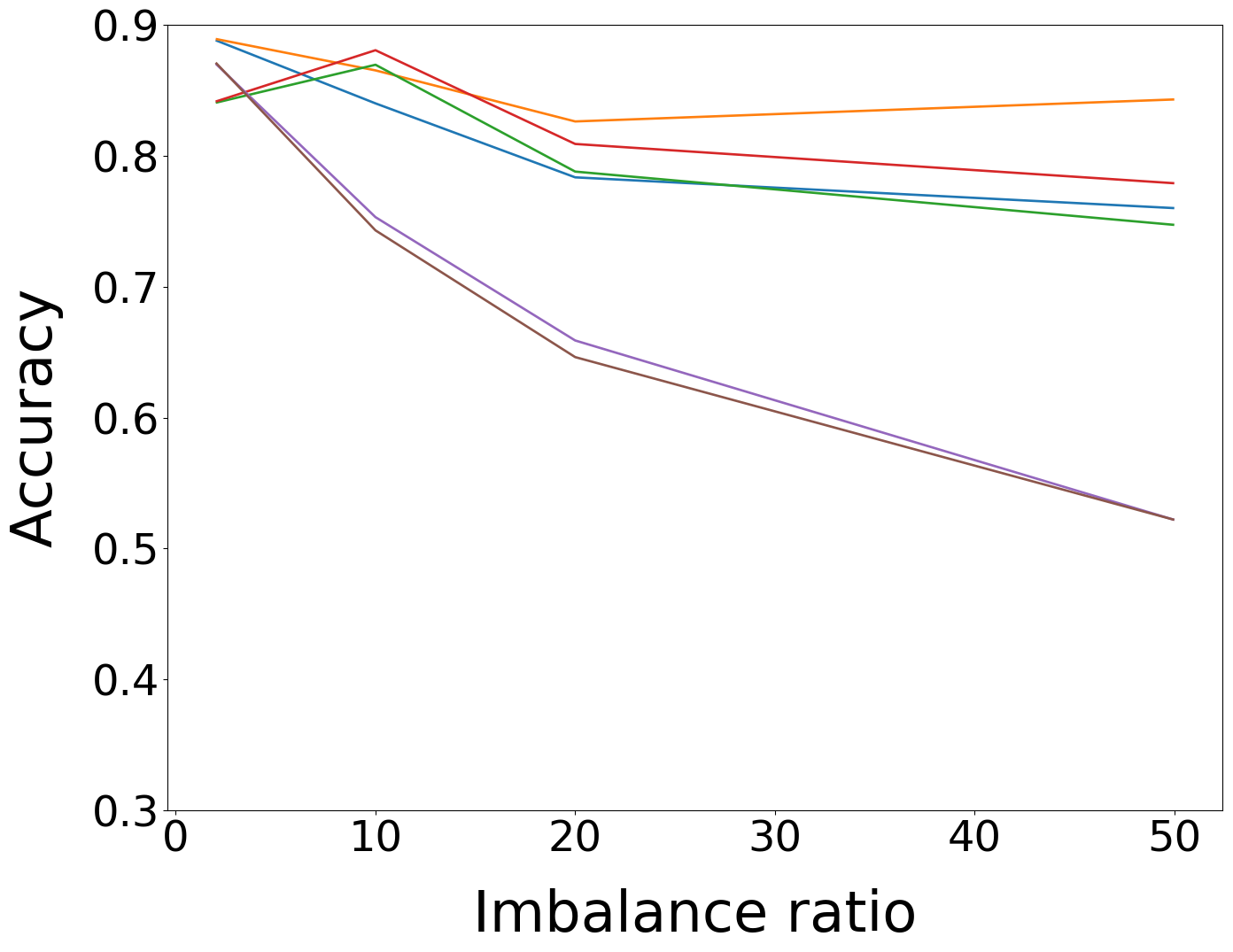}
            \caption{2 minority classes}
            \label{fig:cifar-step_acc_2min}
    \end{subfigure}
    \begin{subfigure}[b]{0.32\textwidth}
            \includegraphics[width=\linewidth]{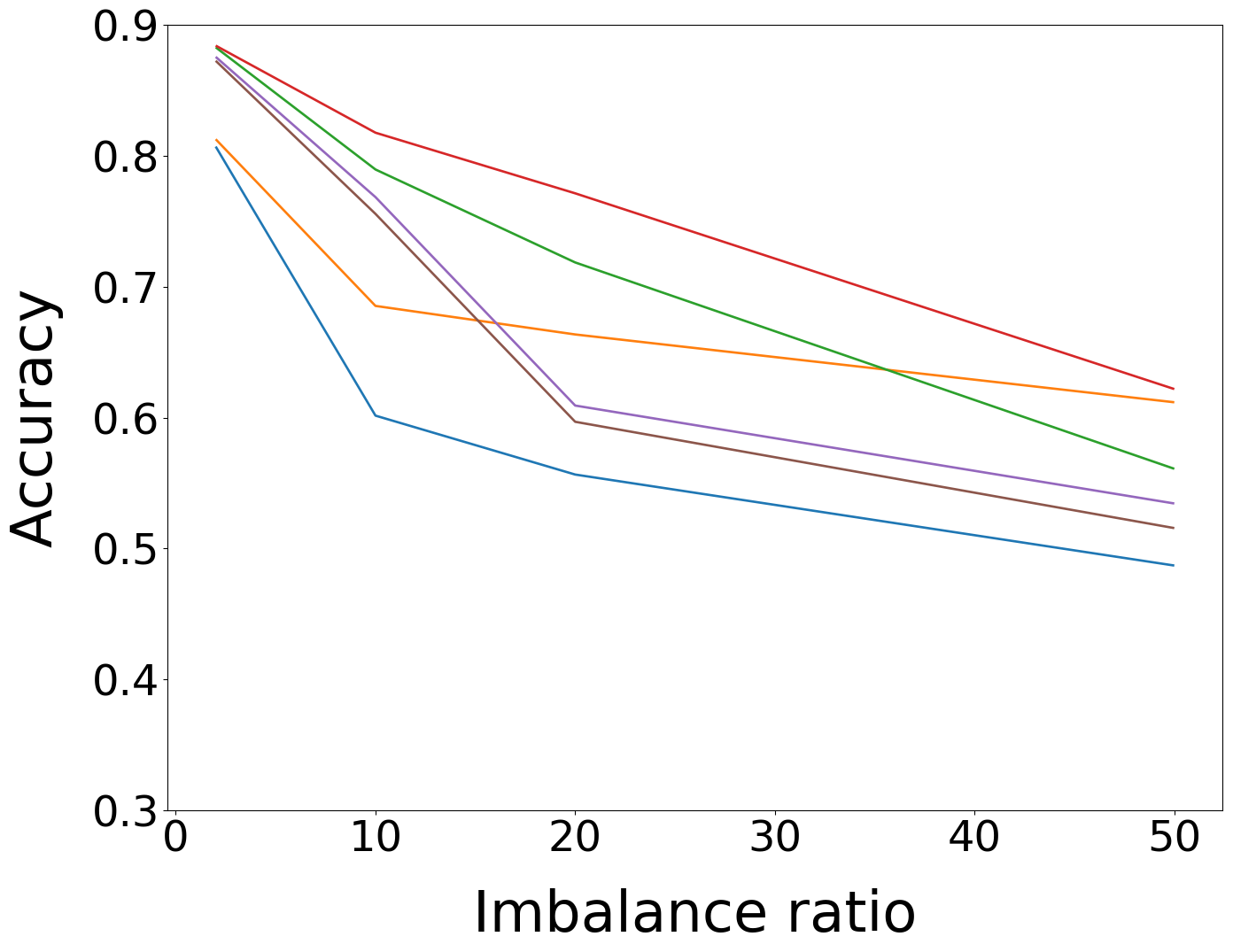}
            \caption{5 minority classes}
            \label{fig:cifar-step_acc_5min}
    \end{subfigure}
    \begin{subfigure}[b]{0.32\textwidth}
            \includegraphics[width=\linewidth]{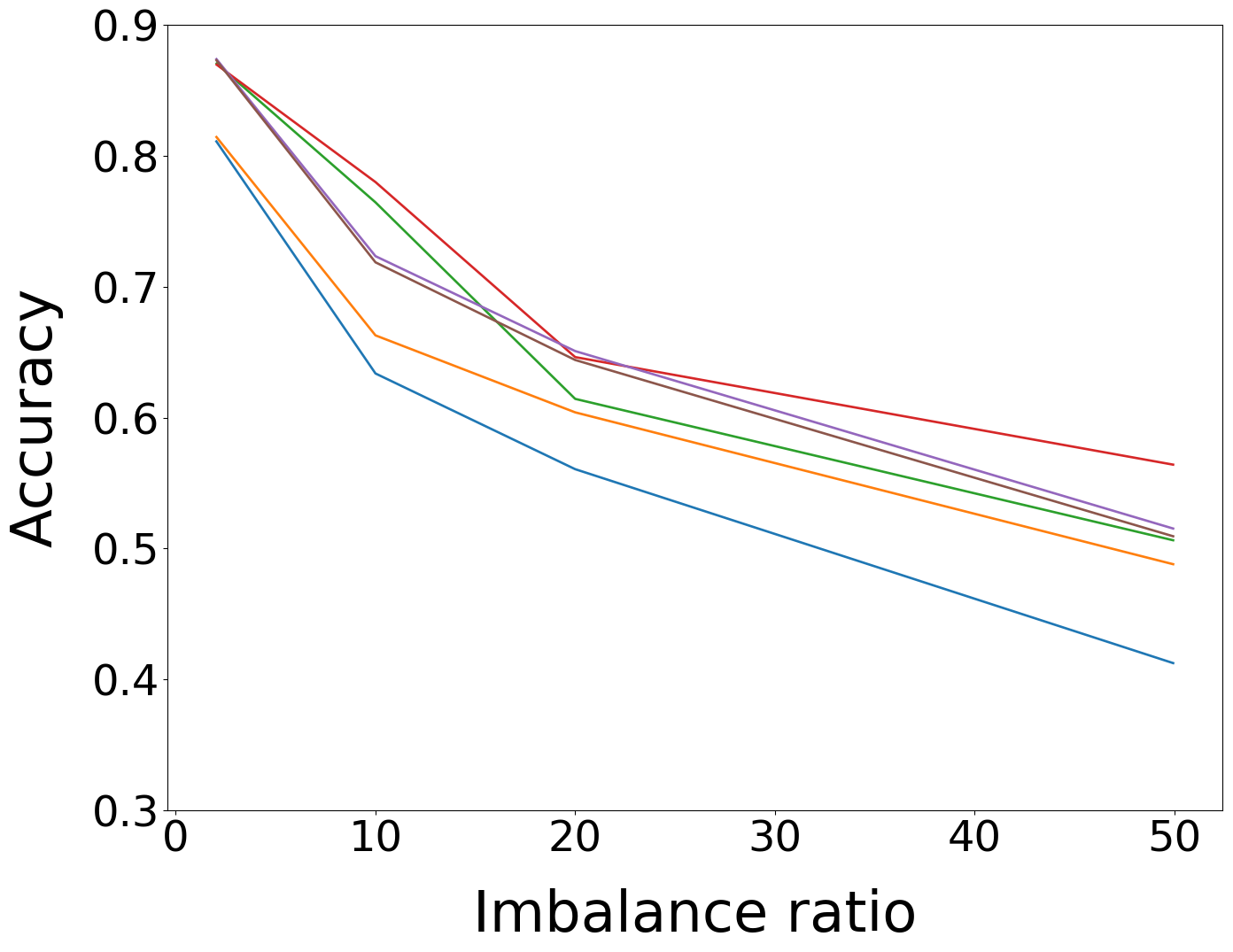}
            \caption{8 minority classes}
            \label{fig:cifar-step_acc_8min}
    \end{subfigure}
    \caption{Comparison of methods with respect to accuracy on MNIST (a~-~c) and CIFAR\nobreakdash-10 (d~-~f) for \textit{step imbalance} with fixed number of minority classes.}
    \label{fig:step_acc_min}
\end{figure}

While our focus is on ROC AUC, we also provide the evaluation of the methods based on overall accuracy measure with results on step imbalance shown in Figure~\ref{fig:step_acc_min}.
As explained in Section~\ref{sec:evaluation}, accuracy has some known limitations and in some scenarios does not reflect the discriminative power of a classifier but rather the prevalence of classes in the training or test set.
Nevertheless, it is still commonly used evaluation score~\cite{haixiang2016learning} and therefore we provide some results according to this metric.

Our results show that thresholding is an appropriate approach to take to offset the prior probabilities of different classes learned by a network based on imbalanced datasets and provided an improvement in overall accuracy.
In general, thresholding worked particularly well when applied jointly with oversampling.

Please note that thresholding does not have an actual effect on the ability of the classifier to discriminate between a given class from another but rather helps to find a threshold on the network output that guarantees a large number of correctly classified cases.
In terms of ROC, multiplying a decision variable by any positive number does not change the area under the ROC curve.
However, finding an optimal operating point on the ROC curve is important when the overall number of correctly classified cases is of interest.

\subsection{Undersampling and oversampling to smaller imbalance ratio}
\label{sec:undersampling}
\begin{figure}[!ht]
    \centering
    \begin{subfigure}[b]{0.45\textwidth}
            \includegraphics[width=\linewidth]{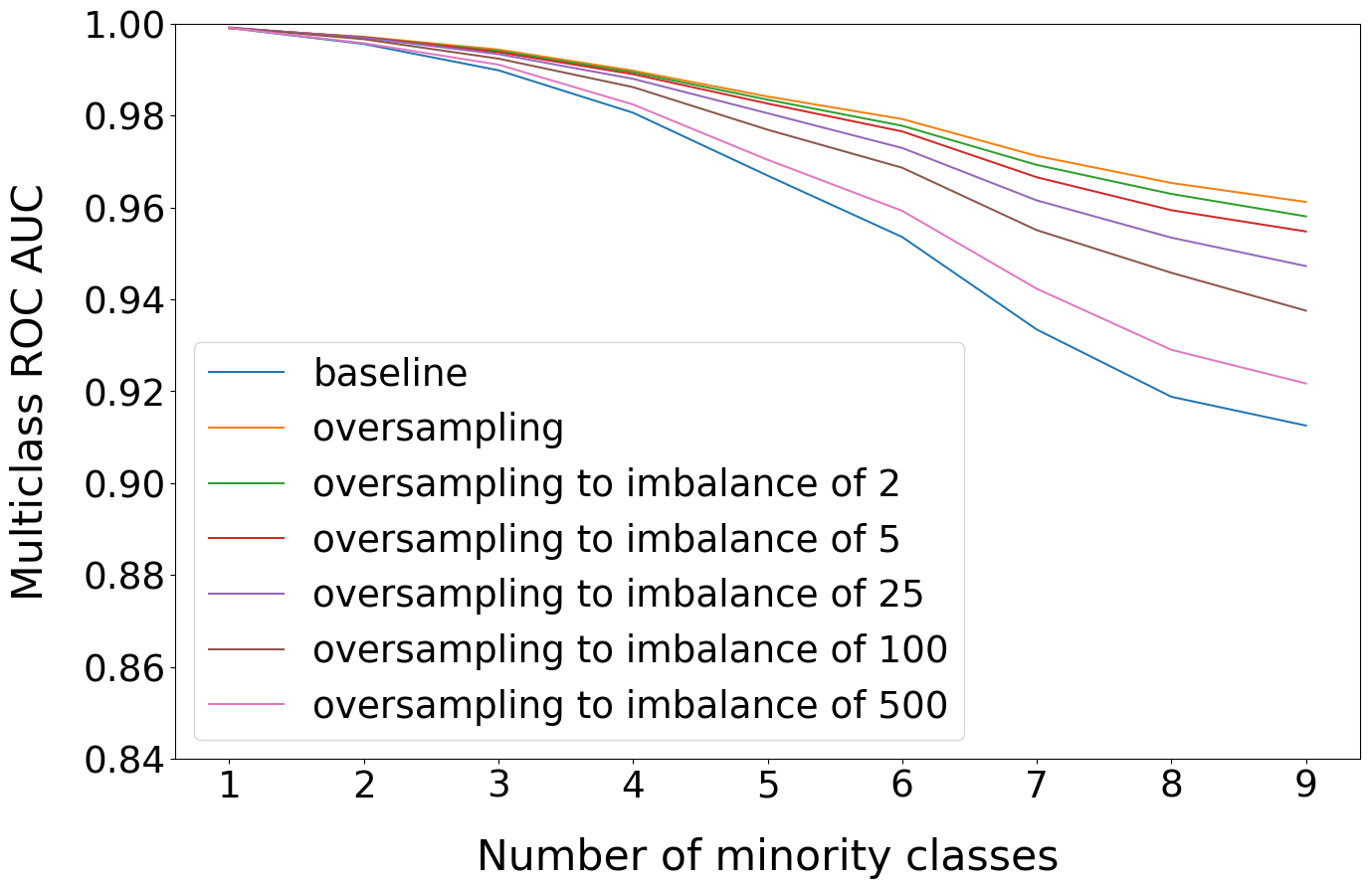}
            \caption{Oversampling}
            \label{fig:reduced_oversampling}
    \end{subfigure}
    \begin{subfigure}[b]{0.45\textwidth}
            \includegraphics[width=\linewidth]{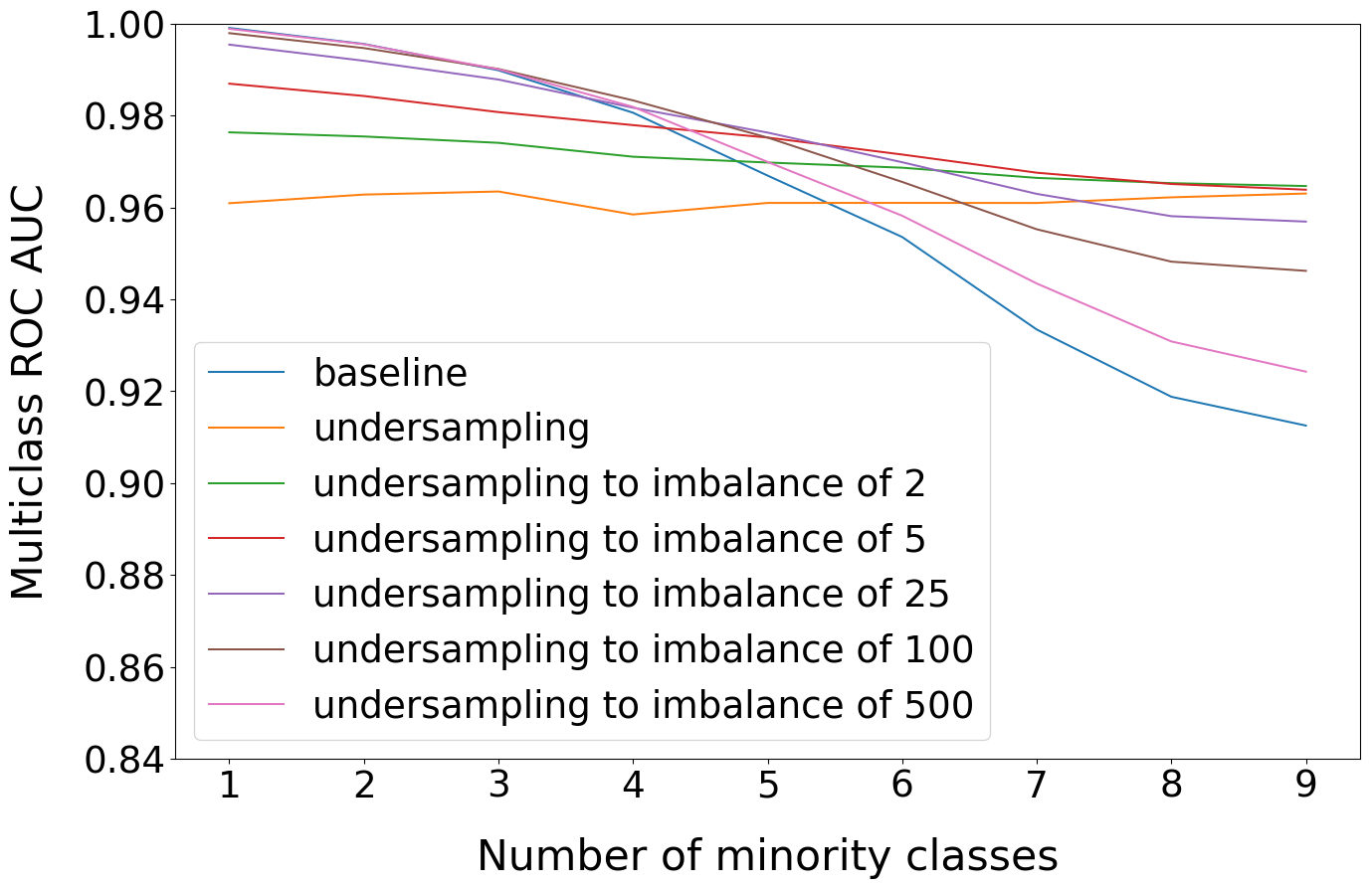}
            \caption{Undersampling}
            \label{fig:reduced_undersampling}
    \end{subfigure}
    \caption{Comparison of oversampling and undersampling to reduced imbalance ratios on MNIST with original imbalance of 1\,000.}
    \label{fig:reduced_sampling}
\end{figure}

The default version of oversampling is to increase the number of cases in the minority classes so that the number matches the majority classes.
Similarly, the default of undersampling is to decrease the number of cases in the majority classes to match the minority classes.
However, a more moderate version of these algorithms could be applied.
For the case of MNIST with imbalance ratio of 1\,000 we have tried to gradually decrease the imbalance with oversampling and undersampling.
The results are shown in Figure~\ref{fig:reduced_sampling}.

The results show that the default version of oversampling was always the best.
Any reduction of imbalance improves the score regardless of the number of minority classes, as shown in Figure~\ref{fig:reduced_oversampling}.
For undersampling, in some cases of moderate number of minority classes, intermediate levels of undersampling performed better than both full undersampling and the baseline.

Moreover, comparing undersampling and oversampling to reduced level of imbalance, we can notice that for each case of oversampling there is a level to which we can apply undersampling and achieve equivalent performance.
However, that level is not known a priori rendering oversampling still the method of choice.

\subsection{Generalization of sampling methods}
\label{sec:generalization}
In some cases undersampling and oversampling perform similarly.
In those cases, one would probably prefer the model that generalizes better.
For classical machine learning models it was shown that oversampling can cause overfitting, especially for minority classes~\cite{chawla2002smote}.
As we repeat small number of examples multiple times, the trained model fits them too well.
Thus, according to this prior knowledge undersampling would be a better choice.
The results from our experiments do not confirm this conclusion for convolutional neural networks.

\begin{figure}[!ht]
    \centering
    \begin{subfigure}[b]{0.31\textwidth}
            \includegraphics[width=\linewidth]{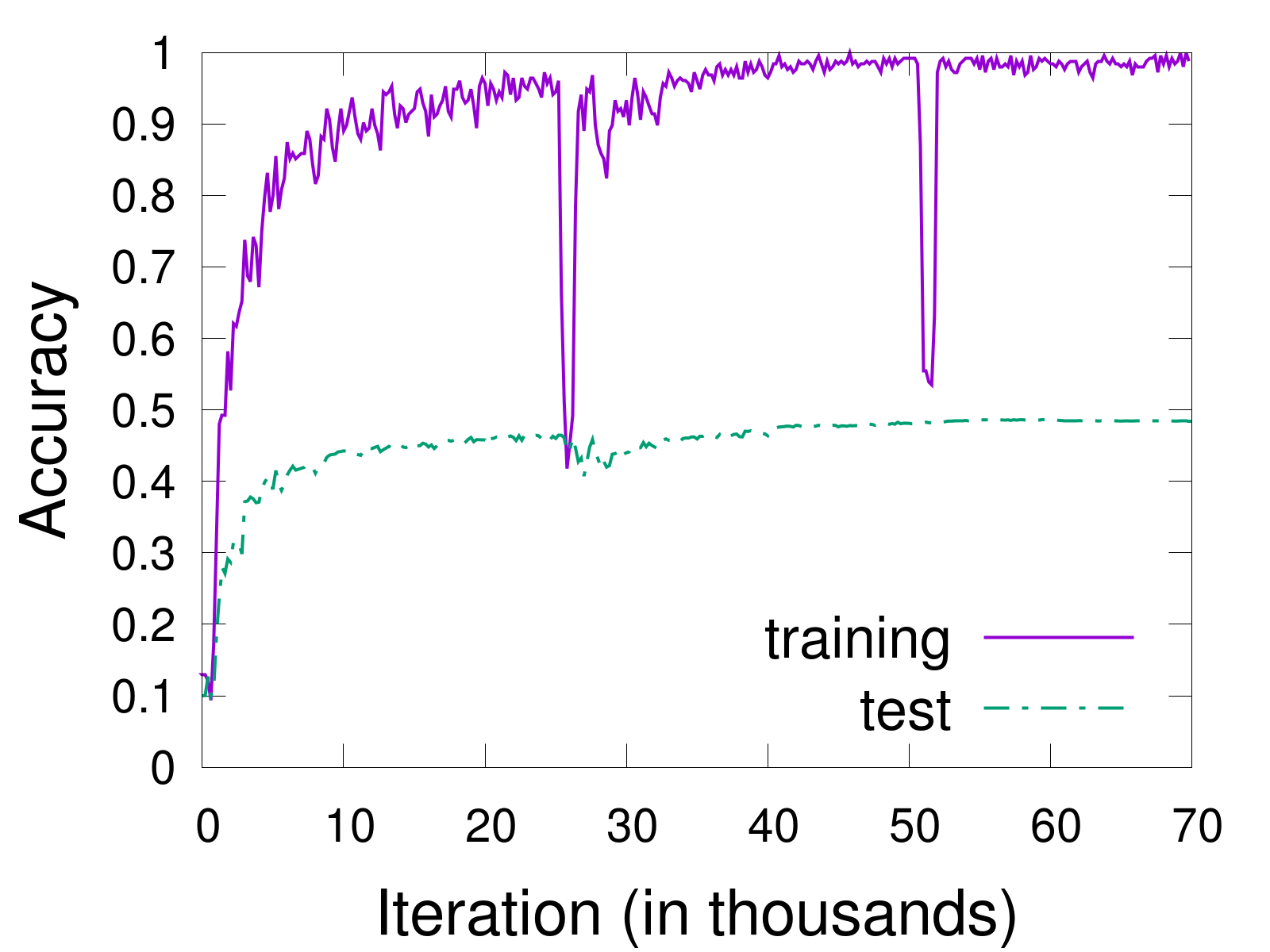}
            \caption{Baseline}
            \label{fig:cifar-convergence-base}
    \end{subfigure}
    \begin{subfigure}[b]{0.31\textwidth}
            \includegraphics[width=\linewidth]{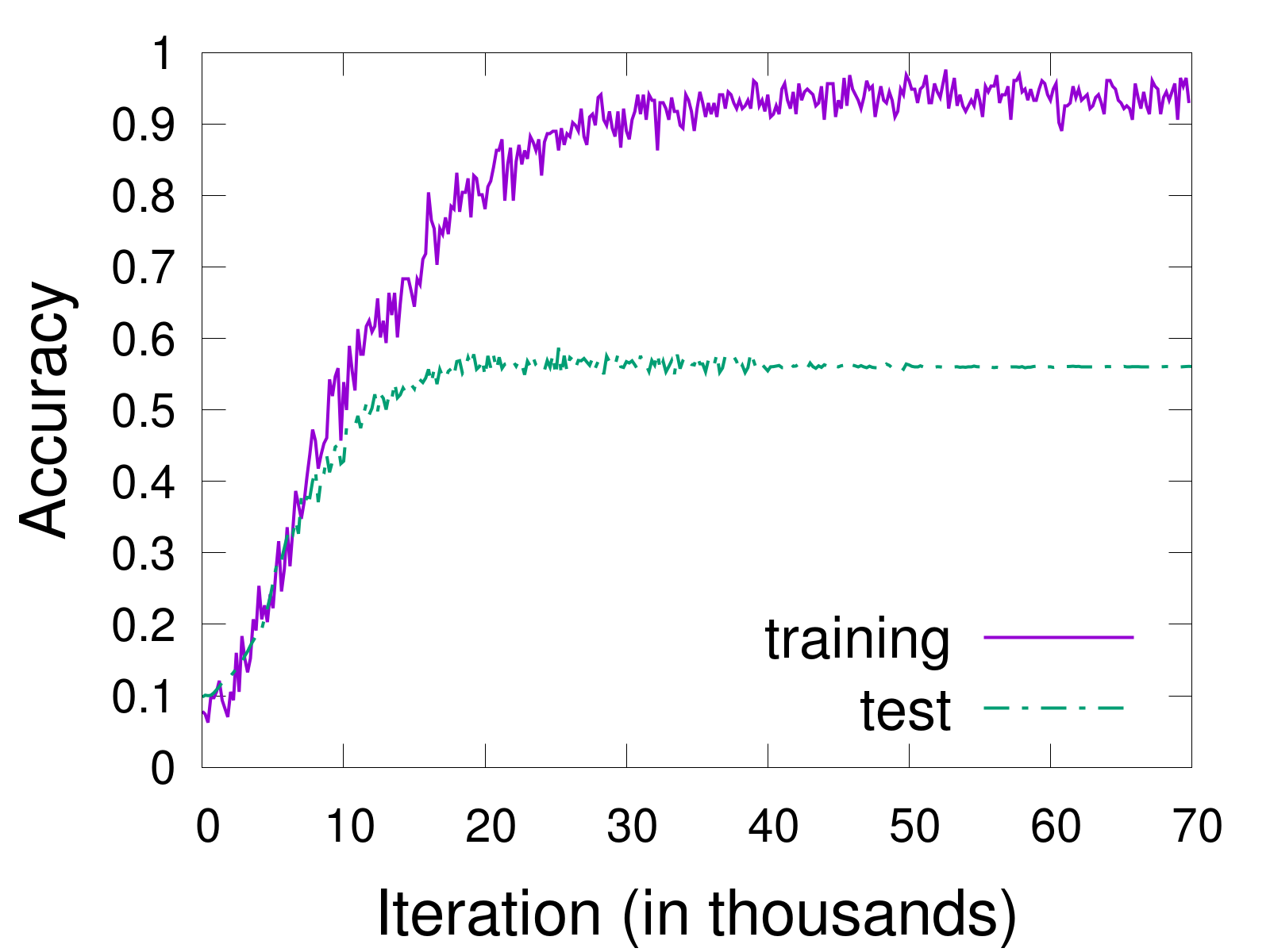}
            \caption{Oversampling}
            \label{fig:cifar-convergence-overs}
    \end{subfigure}
    \begin{subfigure}[b]{0.31\textwidth}
            \includegraphics[width=\linewidth]{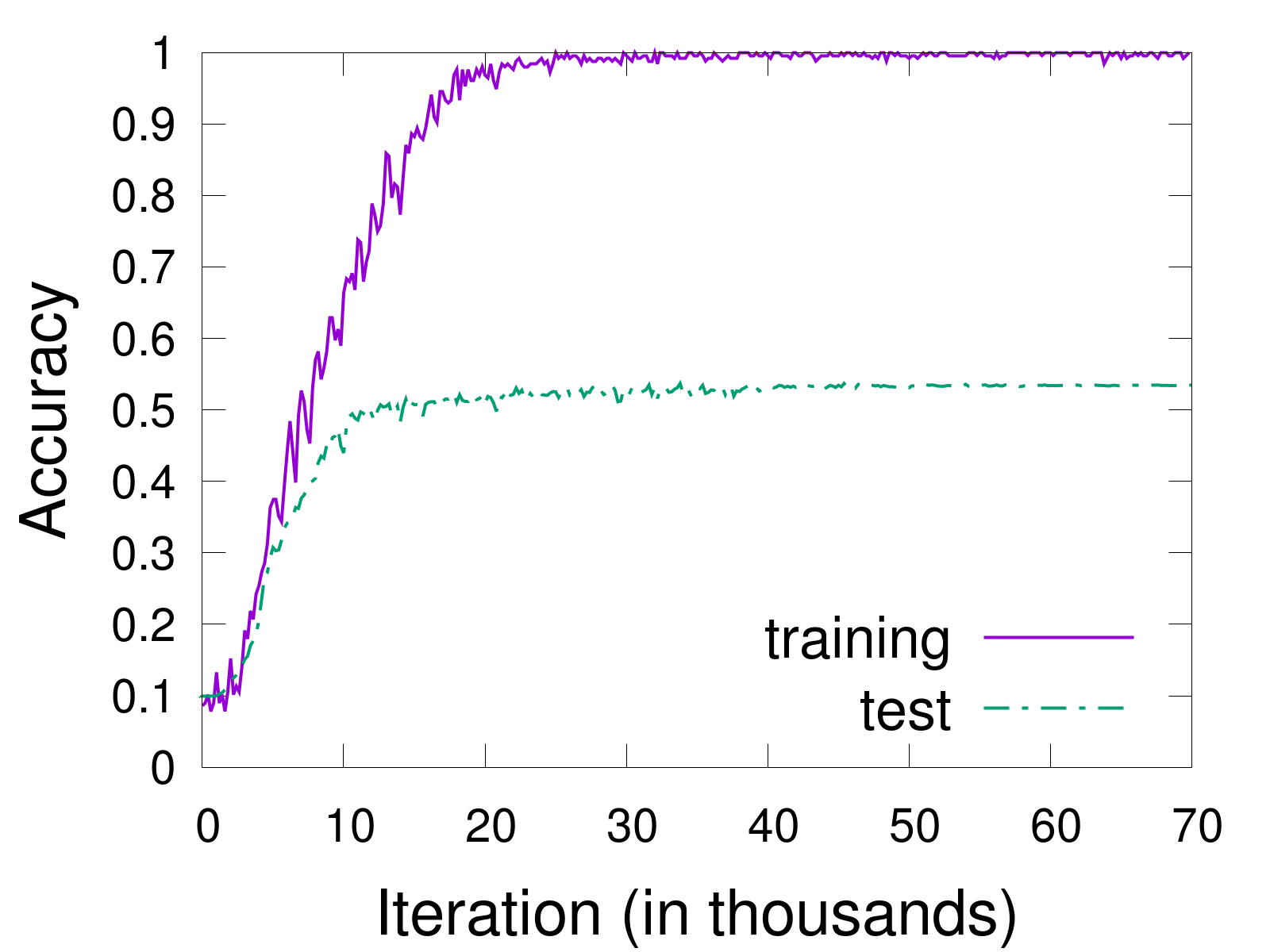}
            \caption{Undersampling}
            \label{fig:cifar-convergence-unders}
    \end{subfigure}
    \caption{Comparison of networks convergence between baseline and sampling methods. Training on CIFAR\nobreakdash-10 \textit{step imbalanced} with 5 minority classes and imbalance ratio of 50.}
    \label{fig:convergence}
\end{figure}

In Figure~\ref{fig:convergence} we compare the convergence of baseline and sampling methods for CIFAR\nobreakdash-10 experiments with respect to accuracy.
Both oversampling and undersampling methods helped to train a better classifier in terms of performance and generalization.
They also made training more stable.
As opposed to traditional machine learning methods, in this case oversampling did not lead to overfitting.
The gap between accuracy on the training and test set does not increase with iterations for oversampling, Figure~\ref{fig:cifar-convergence-overs}.
Furthermore, we validated this phenomenon in multiple additional scenarios for all analyzed datasets and have not observed overfitting in any of these scenarios.
This observation also holds for MNIST and ImageNet datasets and other cases of imbalance.
The additional plots are included in Appendix~A.

\section{Conclusions}
\label{sec:conclusions}

In this study, we examined the impact of class imbalance on classification performance of convolutional neural networks and investigated the effectiveness of different methods of addressing the issue.
We defined and parametrized two representative types of imbalance, i.e. step and linear.
Then we subsampled MNIST, CIFAR\nobreakdash-10 and \nobreak{ImageNet} (ILSVRC\nobreakdash-2012) datasets to make them artificially imbalanced.
We have compared common sampling methods, basic thresholding, and two-phase training.

The conclusions from our experiments related to the class imbalance are as follows.
\begin{itemize}
    \item The effect of class imbalance on classification performance is detrimental.
    \item The influence of imbalance on classification performance increases with the scale of a task.
    \item The impact of imbalance cannot be explained simply by the lower total number of training cases and depends on the distribution of examples among classes.
\end{itemize}

Regarding the choice of a method to handle CNN training on imbalanced dataset we conclude the following.
\begin{itemize}
    \item The method that in most of the cases outperforms all others with respect to multi-class ROC AUC was oversampling.
    \item For extreme ratio of imbalance and large portion of classes being minority, undersampling performs on a par with oversampling.
    If training time is an issue, undersampling is a better choice in such a scenario since it dramatically reduces the size of the training set.
    \item To achieve the best accuracy, one should apply thresholding to compensate for prior class probabilities.
    A combination of thresholding with baseline and oversampling is the most preferable, whereas it should not be combined with undersampling.
    \item Oversampling should be applied to the level that completely eliminates the imbalance, whereas the optimal undersampling ratio depends on the extent of imbalance.
    The higher a fraction of minority classes in the imbalanced training set, the more imbalance ratio should be reduced.
    \item Oversampling does not cause overfitting of convolutional neural networks, as opposed to some classical machine learning models.
\end{itemize}

\section*{Appendix A. Supplementary data}
\label{app:supplementary}

Supplementary material can be found online at \href{https://doi.org/10.1016/j.neunet.2018.07.011}{https://doi.org/10.1016/j.neunet.2018.07.011}.

\clearpage

\bibliography{references}

\noindent\rule{\textwidth}{0.5pt}

\vspace{0.5em}
\noindent
Mateusz Buda, Atsuto Maki, and Maciej A Mazurowski.  A systematic study of the class imbalance problem in convolutional neural networks. (Master’s thesis) Royal Institute of Technology (KTH), 2017. Retrieved from \href{http://urn.kb.se/resolve?urn=urn:nbn:se:kth:diva-219872}{http://urn.kb.se/resolve?urn=urn:nbn:se:kth:diva-219872}

\end{document}